\documentclass[10pt,twocolumn,letterpaper]{article}

\usepackage[pagenumbers]{cvpr} 

\usepackage[binary-units=true]{siunitx}
\usepackage{xcolor}
\usepackage{graphicx}
\usepackage{amsmath}
\usepackage{amssymb}
\usepackage{booktabs}
\usepackage{multirow}
\usepackage{appendix}
\usepackage{float}
\usepackage[acronym]{glossaries}
\usepackage[accsupp]{axessibility}

\usepackage{dblfloatfix}

\usepackage{cuted}

\usepackage[pagebackref,breaklinks,colorlinks]{hyperref}

\usepackage[capitalize]{cleveref}
\crefname{section}{Sec.}{Secs.}
\crefname{table}{Tab.}{Tabs.}

\newacronym{fid}{FID}{Fr\'{e}chet Inception Distance}
\newacronym{fd}{FD}{Fr\'{e}chet distance}
\newacronym{kid}{KID}{Kernel Inception Distance}
\newacronym{gan}{GAN}{Generative Adversarial Networks}
\newacronym{gans}{GANs}{Generative Adversarial Networks}
\newacronym[firstplural=Distributions of Neuron Activations (DNAs)]{dna}{DNA}{Distribution of Neuron Activations}
\newacronym[first=DNA$^\text{hist.}$, shortplural=DNAs$^\text{hist.}$]{dna-h}{DNA$^\text{hist.}$}{Distribution of Neuron Activations (Histograms)}
\newacronym[first=DNA$^\text{Gauss.}$, shortplural=DNAs$^\text{Gauss.}$]{dna-g}{DNA$^\text{Gauss.}$}{Distribution of Neuron Activations (Gaussians)}
\newacronym{emd}{EMD}{Earth Mover's Distance}
\newacronym{nn}{NN}{neural network}
\newacronym{cnn}{CNN}{convolutional neural network}

\def\cvprPaperID{7849} 
\def\confName{CVPR}
\def\confYear{2023}

\makeatletter
\def\blfootnote{\gdef\@thefnmark{}\@footnotetext}
\makeatother

\begin{document}

\title{Visual DNA:\\Representing and Comparing Images using Distributions of Neuron Activations}

\author{Benjamin Ramtoula \qquad  Matthew Gadd \qquad Paul Newman \qquad Daniele De Martini\\
Mobile Robotics Group, University of Oxford\\
{\tt\small \{benjamin, mattgadd, pnewman, daniele\}@robots.ox.ac.uk}
}
\maketitle

\begin{abstract}
	Selecting appropriate datasets is critical in modern computer vision.
	However, no general-purpose tools exist to evaluate the extent to which two datasets differ.
	For this, we propose representing images -- and by extension datasets -- using \glspl{dna}.
	\glspl{dna} fit distributions, such as histograms or Gaussians, to activations of neurons in a pre-trained feature extractor through which we pass the image(s) to represent.
	This extractor is frozen for all datasets, and we rely on its generally expressive power in feature space.
	By comparing two \glspl{dna}, we can evaluate the extent to which two datasets differ with \textit{granular} control over the comparison attributes of interest, providing the ability to customise the way distances are measured to suit the requirements of the task at hand.
	Furthermore, \glspl{dna} are compact, representing datasets of any size with less than 15 megabytes.
	We demonstrate the value of \glspl{dna} by evaluating their applicability on several tasks, including conditional dataset comparison, synthetic image evaluation, and transfer learning, and across diverse datasets, ranging from synthetic cat images to celebrity faces and urban driving scenes.

	\blfootnote{\hspace{-2em}Project page and code:
		\href{https://bramtoula.github.io/vdna/}{\texttt{bramtoula.github.io/vdna}}
	}

\end{abstract}

\glsresetall


\section{Introduction}
\label{sec:intro}

\begin{figure}
	\centering
	\begin{subfigure}[c]{\linewidth}
		\centering
		\includegraphics[width=0.9\textwidth]{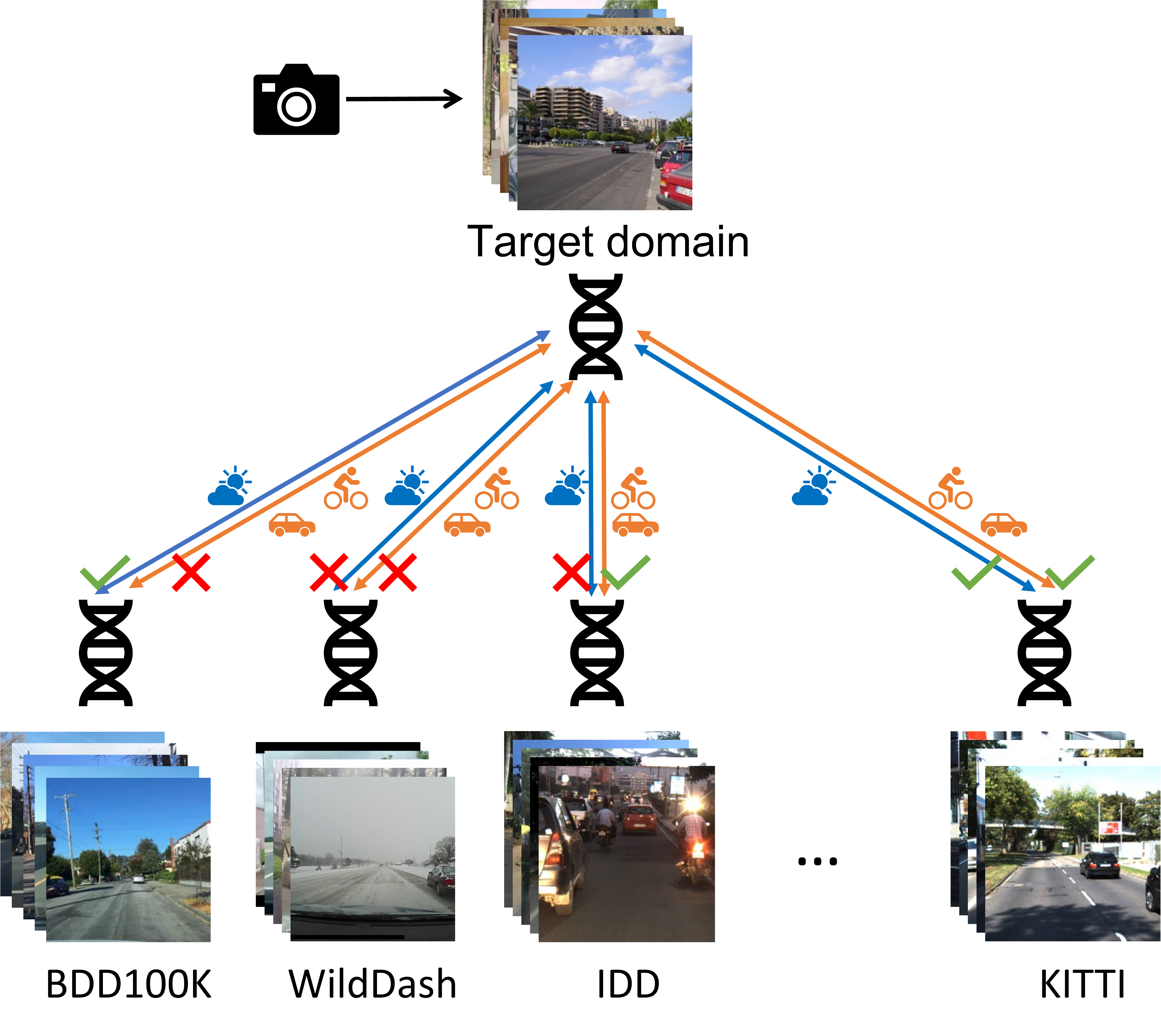}
		\caption{When deploying a vision model in a new target domain, selecting models pre-trained on the most relevant datasets can help. However, there are no methods to measure dataset similarities automatically.
			A general distance between datasets would be sensitive to many variation types, but \acrshortpl{dna} provide sufficient granularity to customise the comparison to focus on features of interest.
			For example, \acrshort{dna} comparisons can be customised to ignore weather conditions or focus on semantic content.}
	\end{subfigure}
	\vspace{5pt}

	\begin{subfigure}[c]{\linewidth}
		\centering
		\includegraphics[width=0.9\textwidth]{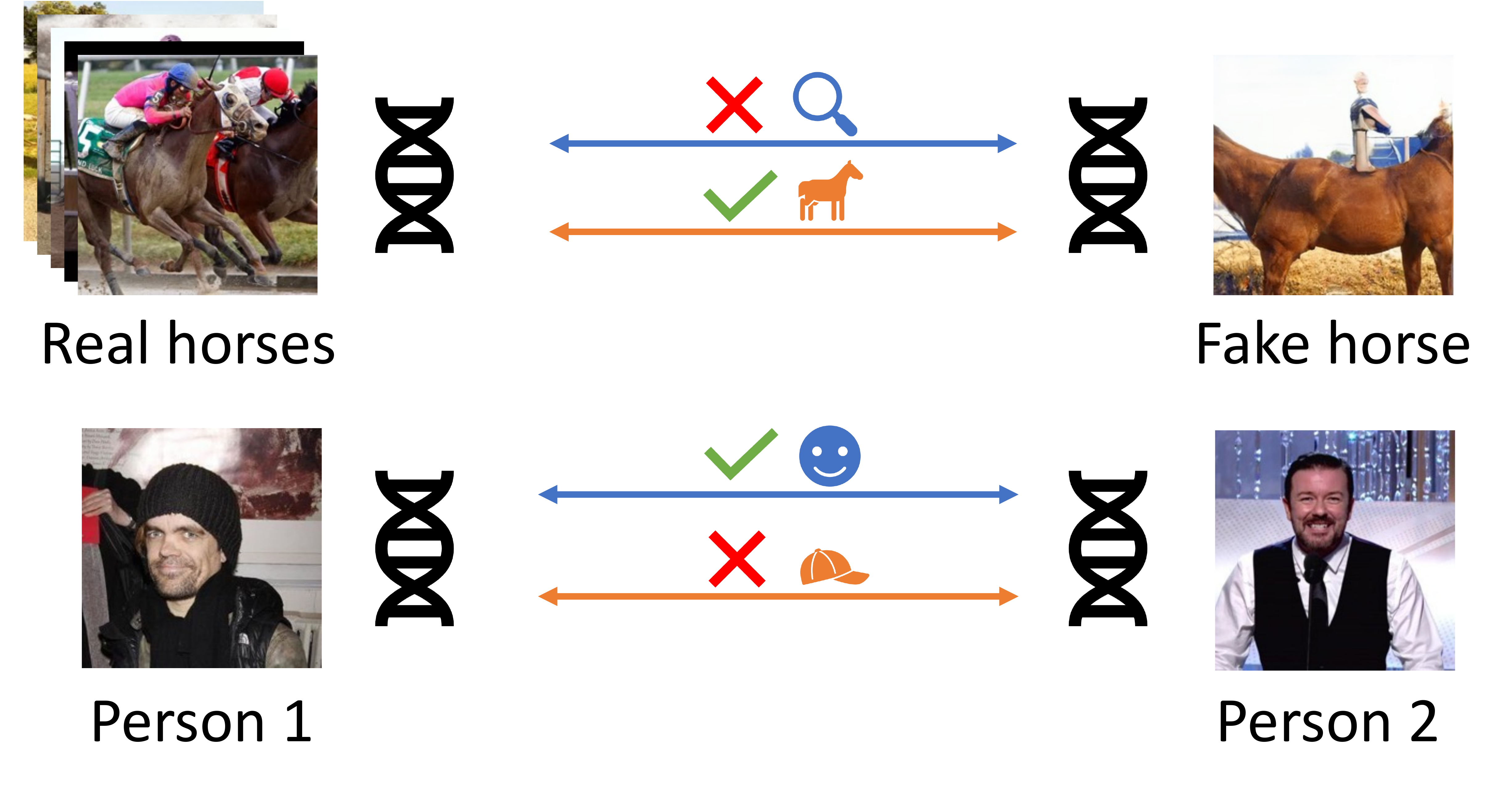}
		\caption{The \acrshort{dna} can also be used to compare individual images to datasets -- for example, to measure the realism and semantic consistency of a synthetic image -- or pairs of images -- for example, to verify the presence of similar attributes such as smiling or wearing a hat.}
	\end{subfigure}
	\vspace{-.15cm}

	\caption{Example use-cases of the \acrshort{dna} representation. \label{fig:meta_comp}}
	\vspace{-.5cm}
\end{figure}

\begin{figure*}[t]
	\centering
	\includegraphics[width=0.8\linewidth]{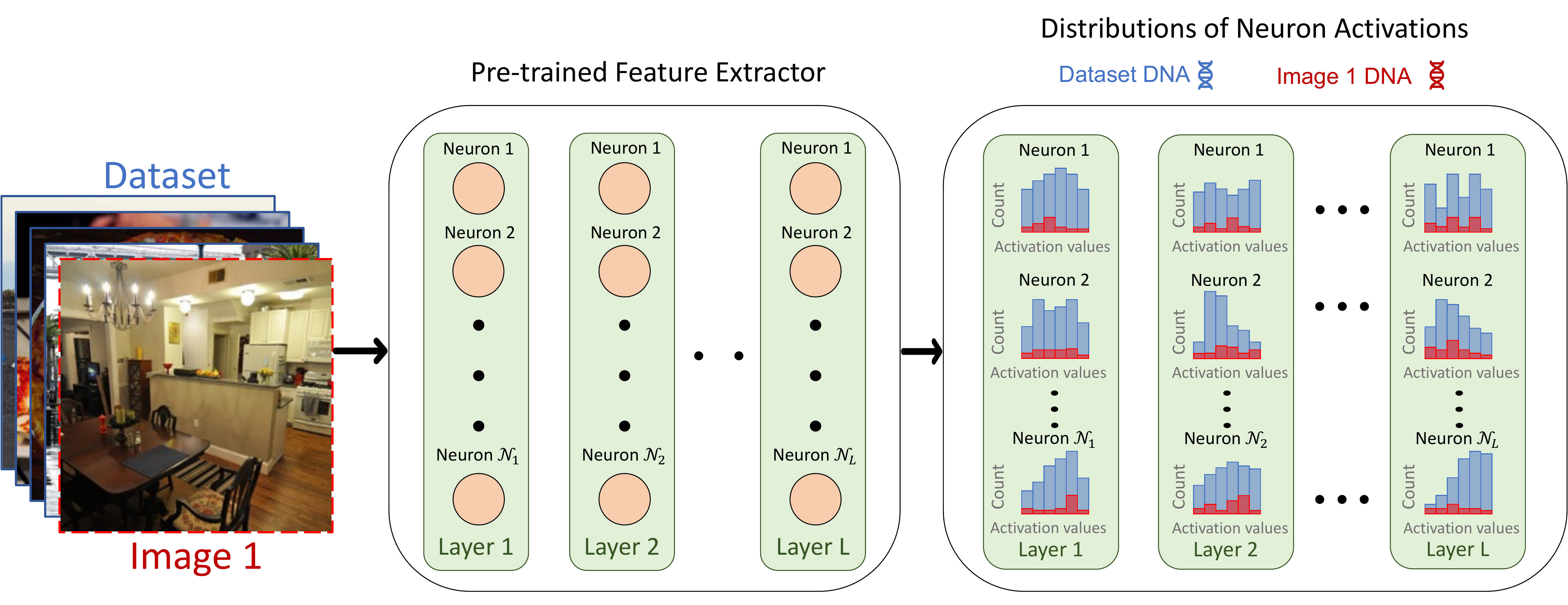}
	\caption{
		We propose representing images by passing them through a pre-trained frozen feature extractor network and collecting neuron activations. We then create a descriptor called the \gls{dna} by fitting a distribution (the histogram in the illustration) to the activations at each neuron.
		We can quantitatively measure the similarity of different datasets or images by comparing their \glspl{dna}. Neuron combination strategies that are sensitive to specific attributes can also allow for customised comparisons of \glspl{dna}.
	}

	\label{fig:dna}
	\vspace{-.5cm}
\end{figure*}

Being able to compare datasets and understanding how they differ is critical for many applications, including
deciding which labelled dataset is best to train a model for deployment in an unlabelled application domain,
sequencing curricula with gradually increasing domain gap,
evaluating the quality of synthesised images, and
curating images to mitigate dataset biases.

However, we currently lack such capabilities.
For example, driving datasets available covering many domains~\cite{yuBDD100KDiverseDriving2020,cordtsCityscapesDatasetSemantic2016b,geigerVisionMeetsRobotics2013,geyerA2D2AudiAutonomous2020,caesarNuScenesMultimodalDataset2020,sunScalabilityPerceptionAutonomous2020,maddernYear1000Km2017,zendelWildDashCreatingHazardAware2018a,huang2018apolloscape,blum2019fishyscapes} were collected under diverse conditions typically affecting image appearance (\eg location, sensor configuration, weather conditions, and post-processing).
Yet, users are limited to coarse or insufficient meta-information to understand these differences.
Moreover, depending on the application, it might be desirable to compare datasets \textit{only} on controlled sets of attributes while ignoring others.
For self-driving, these may be weather, road layout, driving patterns, or other agents' positions.

\glsreset{dna}

We propose representing datasets using their \glspl{dna}, allowing efficient and controllable dataset and image comparisons (\cref{fig:meta_comp}).
The \gls{dna} creation exploits the recent progress in self-supervised representation learning~\cite{goyalScalingBenchmarkingSelfSupervised2019,ericssonSelfSupervisedRepresentationLearning2022} and extracts image descriptors directly from \textit{patterns} of neuron activations in \glspl{nn}.
As illustrated in \cref{fig:dna}, \glspl{dna} are created by passing images through an off-the-shelf pre-trained frozen feature extraction model and fitting a distribution (\eg histogram or Gaussian) to the activations observed at each neuron.
This \gls{dna} representation contains multi-granular feature information and can be compared while controlling attributes of interest, including low-level and high-level information.
Our technique was designed to make comparisons easy, avoiding high-dimensional feature spaces, data-specific tuning of processing algorithms, model training, or any labelling.
Moreover, saving \glspl{dna} requires less than 15 megabytes, allowing users to easily inspect the \gls{dna} of large corpora and compare it to their data before committing resources to a dataset.
We demonstrate the results of using \glspl{dna} on real and synthetic data in multiple tasks, including comparing images to images, images to datasets, and datasets to datasets.
We also demonstrate its value in attribute-based comparisons, synthetic image quality assessment, and cross-dataset generalisation prediction.
\section{Related Works}
\subsection{Studying Image Datasets}
\textbf{Early dataset studies} focused on the limitations of the datasets available at the time.
Ponce \etal~\cite{ponceDatasetIssuesObject2006} highlighted the need for more data, realism, and diversity, focusing on object recognition and qualitative analysis (\eg ``average'' images for each class).
Torralba and Efros~\cite{torralbaUnbiasedLookDataset2011a} found evidence of significant biases in datasets by assessing the ability of a classifier to recognise images from different datasets and measuring cross-dataset generalisation of object classification and detection models. Nowadays, datasets abound, and the approaches used to compare them in those early works would be prohibitive to scale or generalise, often requiring training models for each dataset of interest and access to labels.

\textbf{Compressed datasets representations} allow learning models with comparable properties with reduced dataset sizes.
Dataset distillation approaches~\cite{wangDatasetDistillation2020} \textit{synthesise} a small sample set to approximate the original data when used to train a model.
Core-set selection approaches~\cite{feldmanIntroductionCoresetsUpdated2020}, instead, \textit{select} existing samples, with image-based applications including visual-experience summarisation~\cite{paulVisualPrecisGeneration2014a} and active learning~\cite{senerActiveLearningConvolutional2018}.
While achieving compression of important data properties, these approaches do not produce representations that allow easy dataset comparisons, as our \gls{dna} does.
Modelverse~\cite{lu2022content} performs a content-based search of generative models.
Similarly to \glspl{dna}, they represent multiple datasets -- generated by different generative models -- using distribution statistics of extracted features from the images.
However, their work does not focus on granular and controllable comparisons but on matching a query to the closest distribution.

\textbf{Synthetic data evaluation} for generative models such as \gls{gans}~\cite{borjiProsConsGAN2019} is usually framed as a dataset comparison problem, measuring a distance between datasets of real and fake images.
One of the most widely used metrics is the \gls{fid}~\cite{heuselGANsTrainedTwo2017}, which embeds all images into the feature space of a specific layer of the Inception-v3 network~\cite{szegedyRethinkingInceptionArchitecture2016}.
A multivariate Gaussian is fit to each real and fake embedding, and the \gls{fd} between these distributions is computed.
The \gls{kid}~\cite{binkowskiDemystifyingMMDGANs2018} is another popular approach, which computes the squared maximum mean discrepancy between Inception-v3 embeddings.
There are many other variations, such as using precision and recall metrics for distributions~\cite{djolongaPrecisionRecallCurvesUsing2020,kynkaanniemiImprovedPrecisionRecall2019a,sajjadiAssessingGenerativeModels2018a,simonRevisitingPrecisionRecall2019a}, density and coverage~\cite{naeemReliableFidelityDiversity2020a}, or rarity score~\cite{hanRarityScoreNew2022}.
These approaches rely on high-dimensional features from one layer, while our approach considers neuron activations across layers.
Furthermore, while these measure dataset differences, they have mainly been employed to compare real and synthetic datasets within the same domain, not real ones with significant domain shifts.
Moreover, recent evidence suggests the embeddings typically used can cause a heavy bias towards ImageNet class probabilities~\cite{kynkaanniemiRoleImageNetClasses2022}, motivating more perceptually-uniform distribution metrics.
Additionally, these high-dimensional embeddings make gathering information about specific attributes of interest challenging and lead to computational issues (\eg when clustering).

\subsection{Representation Learning}
\textbf{Feature extractors can provide useful multi-granular features} (\eg containing information about low-level lighting conditions but also the high-level semantics), motivating our design of \glspl{dna}.
Work on the interpretability of \glspl{nn} supports this assumption.
Indeed, Olah \etal~\cite{olahZoomIntroductionCircuits2020, cammarataCurveDetectors2020} explored the idea that \glspl{nn} learn features as fundamental units and that analogous features form across different models and tasks.
Neurons can react more to specific inputs, such as edges or objects~\cite{olahFeatureVisualization2017}.
Combining neuron activations from several images can be a good way to investigate what a network has learned through an activation atlas~\cite{carterActivationAtlas2019}.

\textbf{Existing uses of pre-trained feature extractors} include evaluating computer vision tasks such as Inception-v3 features for \gls{fid} and \gls{kid}, as above. Pre-trained networks on large datasets also provide generally useful representations~\cite{jingSelfsupervisedVisualFeature2019}, which are often fine-tuned for specific applications.
Notably, Evci \etal~\cite{pmlr-v162-evci22a} found that selecting features from subsets of neurons from all layers of a pre-trained network allows better fine-tuning of a classifier head for transfer learning than using only the last layer, suggesting that relevant features are accessible by selecting appropriate neurons.
Moreover, a pre-trained VGG network~\cite{simonyanVeryDeepConvolutional2015} has been used to improve the perceptual quality of synthetic images~\cite{richterEnhancingPhotorealismEnhancement2022,wangHighResolutionImageSynthesis2018a} or to judge photorealism~\cite{zhangUnreasonableEffectivenessDeep2018a, richterEnhancingPhotorealismEnhancement2022}.

\textbf{Self-supervised training} relies on pretext tasks, foregoing labelled data and learning over larger corpora, yielding better representations~\cite{jingSelfsupervisedVisualFeature2019}.
Morozov \etal~\cite{morozovSelfSupervisedImageRepresentations2021} showed that using embeddings from self-supervised networks such as a ResNet50~\cite{heDeepResidualLearning2016} trained with SwAV~\cite{caronUnsupervisedLearningVisual2020} leads to \gls{fid} scores better aligned to human quality judgements when evaluating generative models.
We explore different feature extractors but exploit a ViT-B/16~\cite{dosovitskiyImageWorth16x162021a} trained with Mugs~\cite{zhouMugsMultiGranularSelfSupervised2022} by default, a recent multi-granular feature technique.

\subsection{The need for a more general tool}
\Gls{fid}~\cite{heuselGANsTrainedTwo2017} or \gls{kid}~\cite{binkowskiDemystifyingMMDGANs2018} use representation learning to tackle similar tasks to ours; yet, our formulation extends their applicability.
Quantitative and holistic comparisons between different real datasets have been overlooked, despite being critical to tasks such as transfer or curriculum learning.
We argue that a general data-comparison tool must allow selecting attributes of interest after having extracted a reasonably compact representation of the image(s) and permit the user to \textit{customise} the distance between representations.

\section{DNA - Distributions of Neuron Activations}
Our system is designed around the principle of decomposing images into simple conceptual building blocks that, in combination, constitute uniqueness.
Yet, to cover all possible axes of variations, it is infeasible to specify those building blocks manually.
While we cannot usually link each neuron of a \gls{nn} to a human concept~\cite{olahFeatureVisualization2017}, we show that they provide a useful granular decomposition of images.

Keeping track of neuron activations independently allows us to combine their statistics and study conceptually-meaningful attributes of interest.
As the activations at each neuron are scalar, they can easily be gathered in 1D histograms or univariate Gaussians.
While we would ideally track dependencies between neurons, this is too costly to include in our representation. Nevertheless, we show experimentally that many applications still benefit from \glspl{dna}.

\subsection{Distribution choice}
As in \cref{fig:dna}, in this section, we formulate \glspl{dna} using histograms to fit each neuron's activations distribution. Histograms are a good choice because they do not make assumptions about the underlying distribution; however, we can also consider other distribution approximations.
We also experiment with univariate Gaussians to approximate the activations of each neuron and produce a \gls{dna}, allowing us to describe distributions with only two parameters.
We denote versions using histograms and Gaussians as \gls{dna-h} and \gls{dna-g}, respectively.

\subsection{Generating the DNA from images}
\label{sec:dna-math}
We consider a dataset of images $\mathcal{I}$ where $|\mathcal{I}|~\geq~1$. We also have a pre-trained feature extractor $\mathcal{F}$ with $L$ layers $\mathcal{L}$ manually defined as being of interest, and a set $\mathcal{N}$ of all neurons in those layers.
Each layer $l \in \mathcal{L}$ is composed of $N_l$ neurons, each producing a feature map of spatial dimensions $S_h^l \times S_w^l$.
We can perform a forward pass $\mathcal{F}(i)$ of an image $i \in \mathcal{I}$ and observe the feature map  $f^l_{i}$ obtained at each layer $l$ which has dimensions $N_l~\times~S_h^l~\times~S_w^l$.
For each neuron $n \in \mathcal{N}$ fed with image $i$, we define a histogram $h_i^{n} \in \mathbb{N}^B$ with $B$ pre-defined uniform bins where bin edges are denoted $b_0, b_1,\ldots,b_{B}$.
The count $h^{n}_i[k]$ in the bin of index $k$ for neuron $n$ of layer $l$ is found by accumulating over all spatial dimensions that fall within the bin's edges:
\begingroup
\setlength{\abovedisplayskip}{5pt}
\setlength{\belowdisplayskip}{5pt}
\begin{equation}
	h^{n}_i[k] = \sum^{S_h^l}_{s_h} \sum^{S_w^l}_{s_w}
	\begin{cases}
		1,  & \text{if } f_i^l[n, s_h, s_w]  \in \left[b_k, b_{k+1}\right) \\
		0 , & \text{otherwise}                                             \\
	\end{cases}
\end{equation}
\endgroup
The resulting image's \gls{dna-h} can then be accumulated to represent the dataset $\mathcal{I}$ as $H = \{\mathcal{H}^{n}\}$ for each neuron $n \in \mathcal{N}$, where the element $k$ of $\mathcal{H}^{n}$ can be calculated as:
\begingroup
\begin{equation}
	\mathcal{H}^{n}[k] = \sum_{i \in \mathcal{I}} h^{n}_i[k]
	\label{eq:dna_hist}
\end{equation}
\endgroup

\subsection{Comparing DNAs}
\label{subsec:comp_dna}
Now, comparing \glspl{dna-h} reduces to comparing 1D histograms for neurons of interest.
Depending on the use case, different distances can be considered.
Some tasks might need a distance to be asymmetrical and keep track of original histogram counts, while normalised counts and symmetric distances might be more appropriate for others.

To demonstrate straightforward uses of the representation, we experiment with a widely accepted histogram comparison metric: the \gls{emd}.
Earlier works have argued that the \gls{emd} is a good metric for image retrieval using histograms~\cite{rubnerEarthMoverDistance2000}, with examples of retrieval based on colour and texture.
This work uses its normalised version, equivalent to the Mallows or first Wasserstein distance~\cite{levinaEarthMoverDistance2001}.
The \gls{emd} can be interpreted as the minimum cost of turning one distribution into another, combining the amount of distribution mass to move and the distance.
Specifically, given the normalised, cumulative histogram
\begingroup
\setlength{\abovedisplayskip}{5pt}
\setlength{\belowdisplayskip}{5pt}
\begin{equation}
	\mathcal{F}^{n}[k]=\sum^k_{j=0} \frac{\mathcal{H}^{n}[j]}{\left\lVert\mathcal{H}^{n}\right\lVert_1} \text{ where } k \in 0,\ldots,B-1
\end{equation}
\endgroup
we can easily compute the \gls{emd} between two histograms:
\begingroup
\setlength{\abovedisplayskip}{5pt}
\setlength{\belowdisplayskip}{5pt}
\begin{equation}
	\text{EMD}(\mathcal{H}^{n}_1,\mathcal{H}^{n}_2) = \sum^{B-1}_{k=0} \lvert{\mathcal{F}}^{n}_1[k] - \mathcal{F}^{n}_2[k]\rvert
\end{equation}
\endgroup

As every neuron $n$ is independent in the \gls{emd} formulation, we can vary the contribution of each to the calculation of the total distance.
This allows us to treat neurons differently, \eg when wanting to customise the distance to ignore specific attributes, as presented in \cref{sec:celeba-neuron-selection}.
For this, we introduce the use of a simple linear combination of the histograms through scalar weights $W^{n}$ for each neuron $n$:
\begingroup
\setlength{\abovedisplayskip}{5pt}
\setlength{\belowdisplayskip}{5pt}
\begin{equation}
	\label{eq:total_weights}
	\text{EMD}_W(H_1, H_2) = \sum_{n \in \mathcal{N}} W^{n} {\text{EMD}(\mathcal{H}^{n}_1,\mathcal{H}^{n}_2)}
\end{equation}
\endgroup
In the special case of $W^{n} = {1}/|\mathcal{N}|$, we obtain $\text{EMD}_{\text{avg}}$ as the average \gls{emd} over individual neuron comparisons.

\section{Experimental settings}

\begin{figure}[t]
	\centering
	\includegraphics[width=0.99\linewidth]{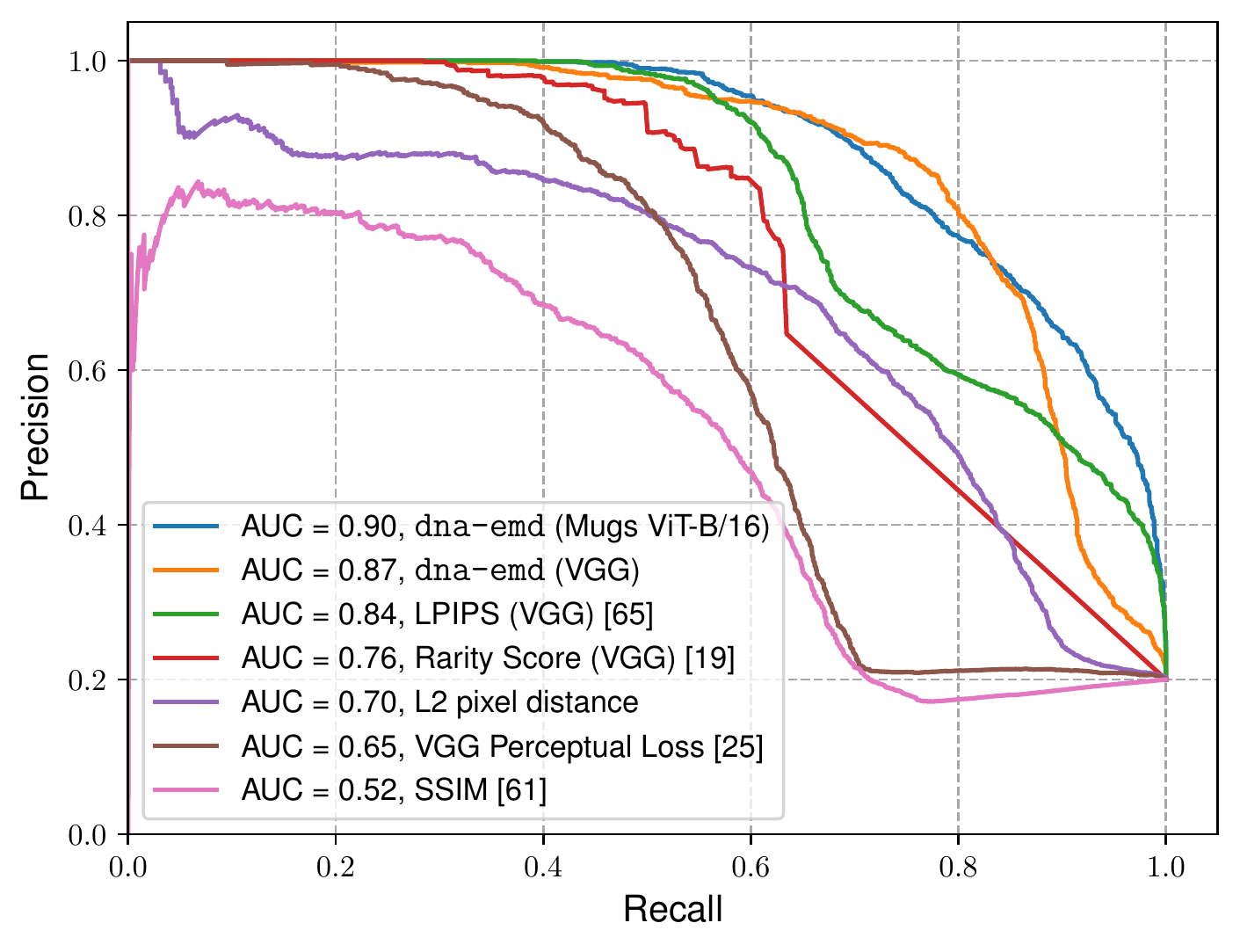}
	\caption{Precision-Recall curves and Area Under the Curve (AUC) for retrieving 2000 individual augmented Cityscapes images mixed with 8000 images from four other datasets. The retrieval compares these images to a reference set of 500 distinct non-augmented Cityscapes images.}
	\label{fig:cs_i2d_quant}
\end{figure}

\begin{figure*}[h!]
	\centering
	\includegraphics[width=0.99\linewidth]{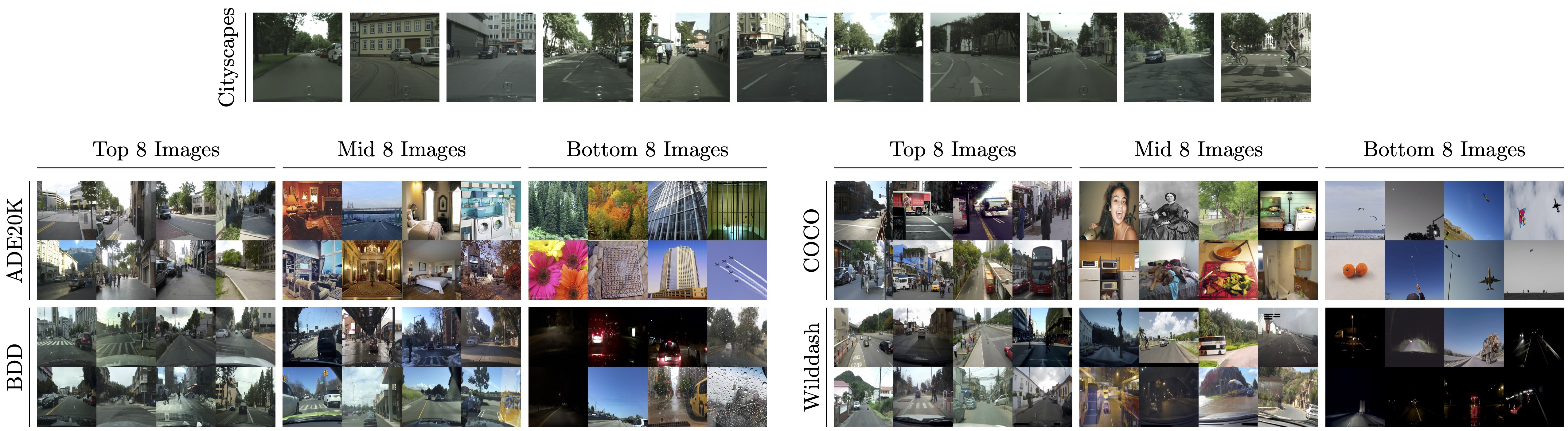}
	\caption{Images from different datasets organised by \texttt{dna-emd} over all neurons against Cityscapes. We successfully discriminate based on the scene type and its visual aspect.
		COCO and ADE20K images get poorly ranked when they do not contain city street scenes.
		The visual aspect and image quality are also considered, as seen with poorly ranked images when Wilddash images have different lighting or BDD images have challenging conditions such as rain on the windshield.
		Images ranked in the middle for BDD tend to contain scenes without any obvious anomaly but with brightness and colours further from Cityscapes images than better-ranked images.}
	\label{fig:cs_i2d}
\end{figure*}

Our experiments evaluate the \gls{dna}'s efficacy on various tasks, datasets, and with diverse feature extractors.

	{\bf Datasets and weights}
We use real and synthetic images from very varied domains -- comparing pairs of images, pairs of datasets, and individual images to datasets.
Images are processed using tools provided by Parmar \etal~\cite{parmarAliasedResizingSurprising2022}.
Notably, no additional tuning is performed for any experiments, i.e. \textit{the feature extractors' weights are frozen}.

{\bf Settings}
\label{sec:settings}
Activation ranges vary for different neurons; it is therefore important to adjust the histogram settings for each neuron to get balanced distances between neurons.
We thus monitor each neuron's activation values over a large set of datasets and track the minimum and maximum values observed, adding a margin of $20\%$ and using these extremes to normalise activations between $-1$ and $1$.
Notably, the \textit{only} hyperparameter for the \gls{dna-h} is the number of bins, $B$, which we set to 1000 for our experiments.

	{\bf Benchmarking}
Our primary baseline, \texttt{fd}, is the \acrlong{fd}~\cite{heuselGANsTrainedTwo2017}, which measures the distance of two multivariate Gaussians that fit the samples in the embedding space of entire layers of the extractor. We use the acronym ``\acrfull{fd}'' rather than ``\acrfull{fid}'' as we explore different feature extractors than Inception-v3.
Traditionally, \texttt{fd} has been used on a single layer, but we show its performance on different combinations of the extractor layers.
Here, \texttt{dna-emd} denotes \gls{emd} comparisons of our \glspl{dna-h}, and \texttt{dna-fd} denotes \gls{fd} comparisons of our \glspl{dna-g}.
These three settings allow us to verify our approach, showing the effectiveness of considering every neuron as independent \textit{and} not constraining the activations to fit specific distributions.

We do not consider the Kernel Distance~\cite{binkowskiDemystifyingMMDGANs2018} as it is
unclear how to compress the required information in a compact representation.

	{\bf Memory footprint} We provide details on memory complexity and \gls{dna} storage size in \Cref{tab:mem_compl}.

\begin{table}[htb]
	\centering
	\renewcommand{\arraystretch}{1.0}
	\resizebox{1.0\columnwidth}{!}{
		\begin{tabular}{lcccc}
			Method                      & Complexity                   & Theoretical size      & Observed size          \\
			\hline
			Features                    & $N \times n \times S$ floats & \SI{1.10}{\tera\byte} & -                      \\
			Spatially averaged features & $N \times n$ floats          & \SI{5.59}{\giga\byte} & \SI{2.91} {\giga\byte} \\
			\gls{dna-g}                 & $2 \times n$ floats          & \SI{159}{\kilo\byte}  & \SI{84.7}{\kilo\byte}  \\
			\gls{dna-h}                 & $B \times n$ ints            & \SI{79.9}{\mega\byte} & \SI{14.8}{\mega\byte}  \\
		\end{tabular}
	}
	\caption{Memory footprint of data for
		$N$ images, $n$ feature extractor neurons with an average of $S$ elements in their feature maps, and $B$ bins.
		Example with FFHQ (\SI{89.1}{\giga\byte}, $N$$=$$70\text{k}$), Mugs ViT-B/16 ($n$$=$$9984$, $S$$=$$197$) and $B$$=$$1000$ bins. Observed sizes are from files saved using NumPy's \texttt{savez\_compressed} function.}
	\label{tab:mem_compl}
\end{table}

\section{Results}

\subsection{Finding most similar images with domain shifts}
\label{subsec:cs-retrieval}

We first show the ability of \glspl{dna} to find real images similar to a reference dataset.
We have created two datasets: a \textit{reference} $D_r$ contains $500$ random Cityscapes~\cite{cordtsCityscapesDatasetSemantic2016b} images; a \textit{comparison} $D_c$ contains 2000 images from each of ADE20K~\cite{zhouSceneParsingADE20K2017}, BDD~\cite{yuBDD100KDiverseDriving2020}, COCO~\cite{linMicrosoftCOCOCommon2014a}, Wilddash~\cite{zendelWildDashCreatingHazardAware2018a} as well as 2000 randomly augmented images (\eg noise, blur, spatial and photometric) from Cityscapes that are not present in $D_r$.
We rank each image from $D_c$ in terms of its distance to $D_r$ as a whole.
We expect the top-ranked $D_c$ images to all be Cityscapes augmentations.
We compare the use of \texttt{dna-emd} using the $\text{EMD}_\text{avg}$ (\cref{subsec:comp_dna}) to other perceptual comparison baselines: a perceptual loss~\cite{vgg}, LPIPS~\cite{zhangUnreasonableEffectivenessDeep2018a}, SSIM~\cite{ssim}, the L2 pixel distance, and rarity score~\cite{hanRarityScoreNew2022}.
All approaches are evaluated using features from VGG~\cite{simonyanVeryDeepConvolutional2015}.
For approaches comparing image pairs (all except ours \& rarity score), we define the distance for one $D_c$ image to $D_r$ as its average distance to each image in $D_r$.

\cref{fig:cs_i2d_quant} shows \glspl{dna} performing best at this task \textit{while not requiring expensive pairwise image comparisons}. Results are further improved when using features from a self-supervised approach, Mugs~\cite{zhouMugsMultiGranularSelfSupervised2022}, instead of VGG.

We also verify qualitatively how images from different datasets from $D_c$ are ranked when compared to $D_r$ using Mugs features.
\cref{fig:cs_i2d} shows the successful discrimination of scene types and visual aspects in all comparisons.

\begin{figure}[ht]
	\centering
	\begin{subfigure}[b]{0.95\linewidth}
		\centering
		\includegraphics[width=\linewidth]{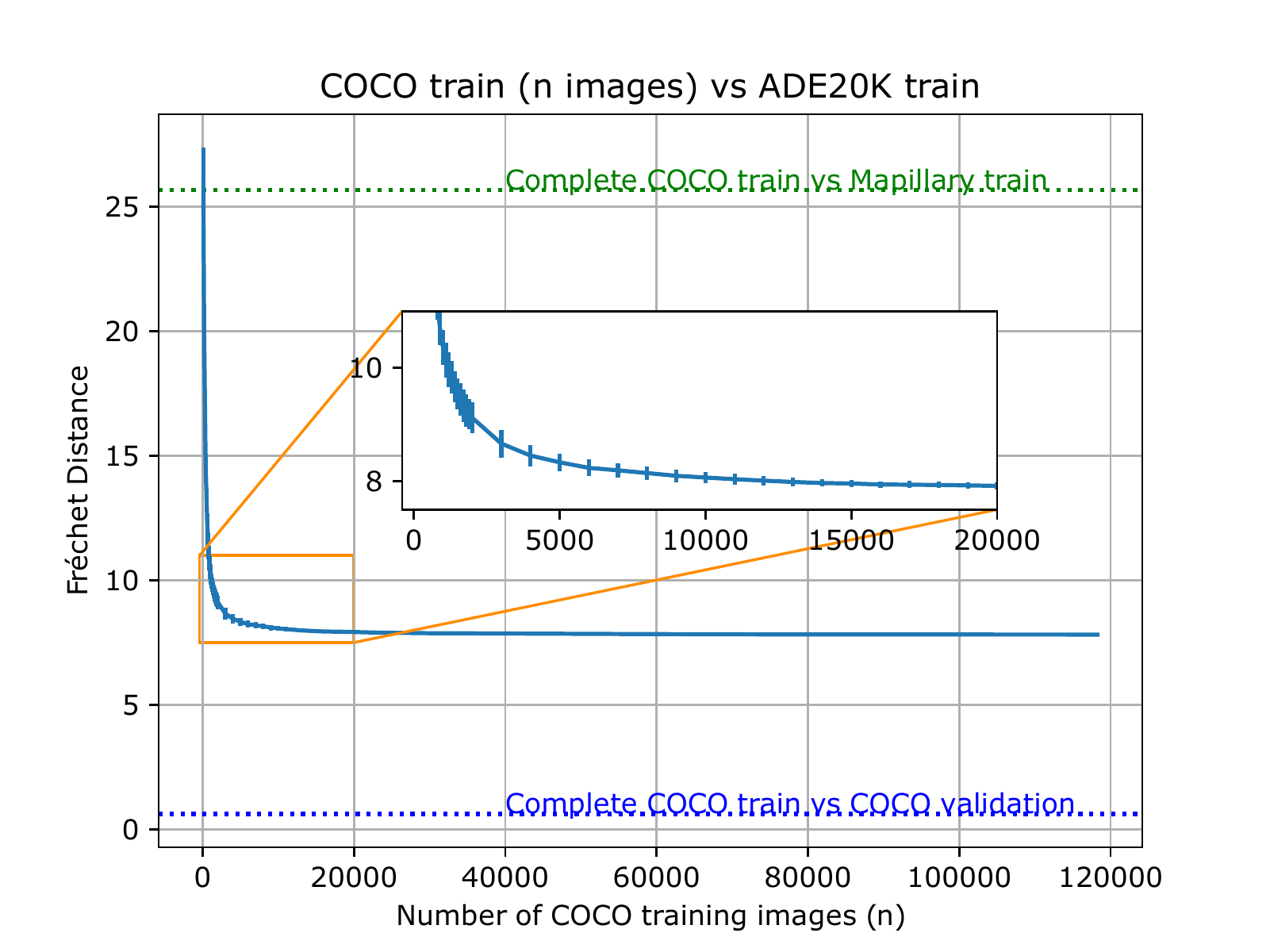}
		\caption{Fr\'echet Distance (\texttt{fd})}
	\end{subfigure}

	\begin{subfigure}[b]{0.95\linewidth}
		\centering
		\includegraphics[width=\linewidth]{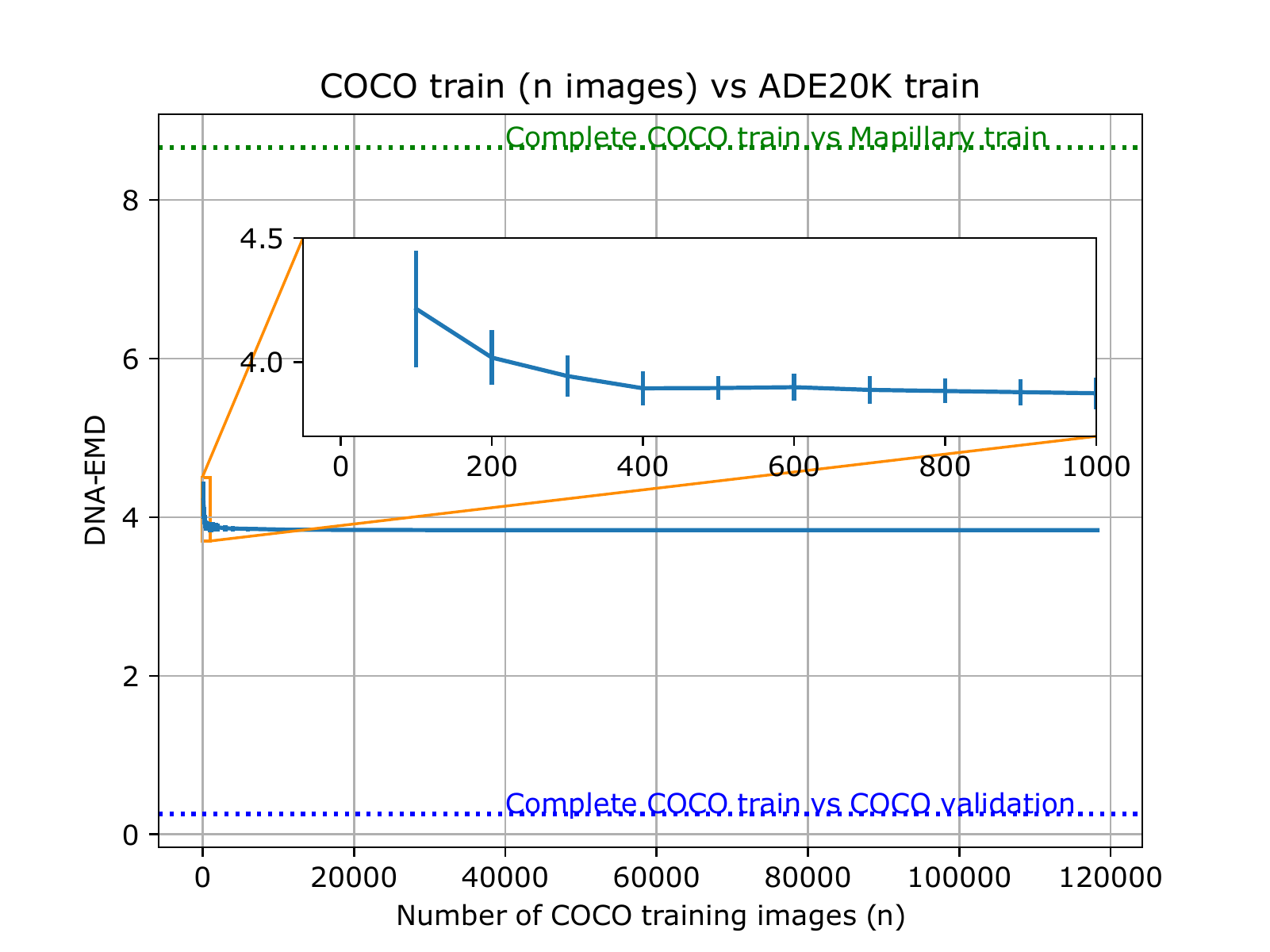}
		\caption{Ours (\texttt{dna-emd})}
	\end{subfigure}
	\caption{Influence of the number of images on dataset distances using DINO (ViT-B/16) features. \texttt{dna-emd} requires significantly fewer samples than \texttt{fd} from the COCO training set to reach a steady value when compared with ADE20K's training set.
		Here, the distances using all images are non-zero, as there is a domain shift between the datasets. Dashed lines illustrate the distances obtained comparing COCO's training set to its validation set and to ADE20K to illustrate the scale of the error. Results are averaged over ten seeds, with vertical lines showing the standard deviations.}
	\label{fig:bias}
	\vspace{-0.5cm}
\end{figure}

\subsection{Number of images required for dataset DNAs}

\texttt{fd} is known to work poorly with scarce data~\cite{binkowskiDemystifyingMMDGANs2018}.
We assess this in~\cref{fig:bias}, comparing the distance between the entire ADE20K training set~\cite{zhouSceneParsingADE20K2017} and increasingly larger subsets of COCO's training set~\cite{linMicrosoftCOCOCommon2014a}.
Here, \texttt{fd} reaches a steady value after 10000 samples while \texttt{dna-emd} needs only 400, making it a reliable representation even for small datasets.

\subsection{Ignoring specific attributes}
\label{sec:celeba-neuron-selection}

Here, we demonstrate the granularity provided by individual neurons by considering: given \glspl{dna} of two datasets, can we measure the distance between them while ignoring contributions due to specifically-selected attributes?

\textbf{Attribute datasets} For this experiment, we split the CelebA training set according to each of the 40 labelled attributes, \eg \textit{smiling} or \textit{wearing a hat}, where $A$ is the set of all attributes. We use the ``in the wild'' version of images which are not cropped and aligned around faces, allowing us to assess robustness to different locations and scales of attributes.
Considering one attribute $a \in A$, we compute two \glspl{dna}, $\mathcal{D}_a$, with images \textit{with} the attribute, and $\mathcal{D}_{\bar{a}}$, with images \textit{without} the attribute.
Neurons whose distributions vary greatly between these \glspl{dna} -- \ie are \textit{sensitive} -- correlate with the attribute.

\textbf{Learned sensitivity removal and deviation} We input the neuron-wise (for \texttt{dna-fd} and \texttt{dna-emd}) or layer-wise (for \texttt{fd}) distances between $\mathcal{D}_a$ and $\mathcal{D}_{\bar{a}}$ into a linear layer, which produces a weighted distance with which we can ignore differences of a specific attribute while maintaining sensitivity to the other attributes.
Its parameters correspond to the weights $W$ for the linear combination in~\cref{eq:total_weights}.
Next, we define the \textit{sensitivity deviation} of attribute $a$. For \texttt{dna-emd}:
\begingroup
\setlength{\abovedisplayskip}{5pt}
\setlength{\belowdisplayskip}{5pt}
\begin{equation}\label{eqn:sremov}
	\Delta_a = 1 - \frac{\text{EMD}_W(\mathcal{D}_a, \mathcal{D}_{\bar{a}})}{\text{EMD}_{\text{avg}}(\mathcal{D}_a, \mathcal{D}_{\bar{a}})}
\end{equation}
\endgroup
This and all the following calculations can be applied to \texttt{fd} and \texttt{dna-fd} with the \gls{fd}.
If $a$ is the only attribute that changes between $\mathcal{D}_a$ and $\mathcal{D}_{\bar{a}}$, and $W$ is optimised to ignore $a$, then the $\text{EMD}_W$ should not be sensitive to $a$ and $\text{EMD}_W(\mathcal{D}_a, \mathcal{D}_{\bar{a}}) = 0$, $\Delta_a = 1$.
For instance, we have datasets at night and datasets at day but want to compare only considering the types of vehicles present.
For attributes to which we want the distance to remain sensitive, $b \in A \setminus a$, we can also measure deviations from the original distance caused by the weights $W$ using $|\Delta_b|$, indicating the change in sensitivity of the distance to this attribute. We want no deviation for these attributes, \ie $|\Delta_b| = 0$.
Finally, we impose (and back-propagate, using Adam~\cite{kingma2014adam}) a loss:
\begingroup
\setlength{\abovedisplayskip}{5pt}
\setlength{\belowdisplayskip}{5pt}
\begin{equation}
	\label{eq:loss}
	L_a = | 1 - \Delta_a | + \frac{1}{|A| - 1} \sum_{b\in{}A\setminus{}a} | \Delta_b |
\end{equation}
\endgroup
meaning that we will optimise $W$ to desensitize the $\text{EMD}_W$ to $a$ but remain sensitive to all other attributes in $b \in A \setminus a$.

	{\bf CelebA sensitivities}
\Cref{tab:celeba-atts} presents results as averaged over all attributes (\ie with $a=\mathtt{hat}$ and $b$ being all other attributes, then $a=\mathtt{glasses}$ and $b$ all others, etc.).
The results clearly show that neuron granularity is crucial for success as \texttt{fd}, which operates layer-wise, falls short against \texttt{dna-emd} and \texttt{dna-fd}.
Averaged over all attributes, our approach can discard $95.5\%$ of the distance over the attributes on which we remove sensitivity, while only causing a $9.6\%$ of deviation in distances over other attributes.
\texttt{dna-emd} performs slightly better than \texttt{dna-fd}, but both do very well.
We observe that all feature extractors considered can somewhat succeed at the task, including the ResNet-50 with random weights, which we expect to still produce valuable features~\cite{ramanujanWhatHiddenRandomly2020}.
However, we obtain the best results using self-supervised models which are likely to produce more informative features.

\begin{figure}
	\centering
	\includegraphics[width=0.95\linewidth]{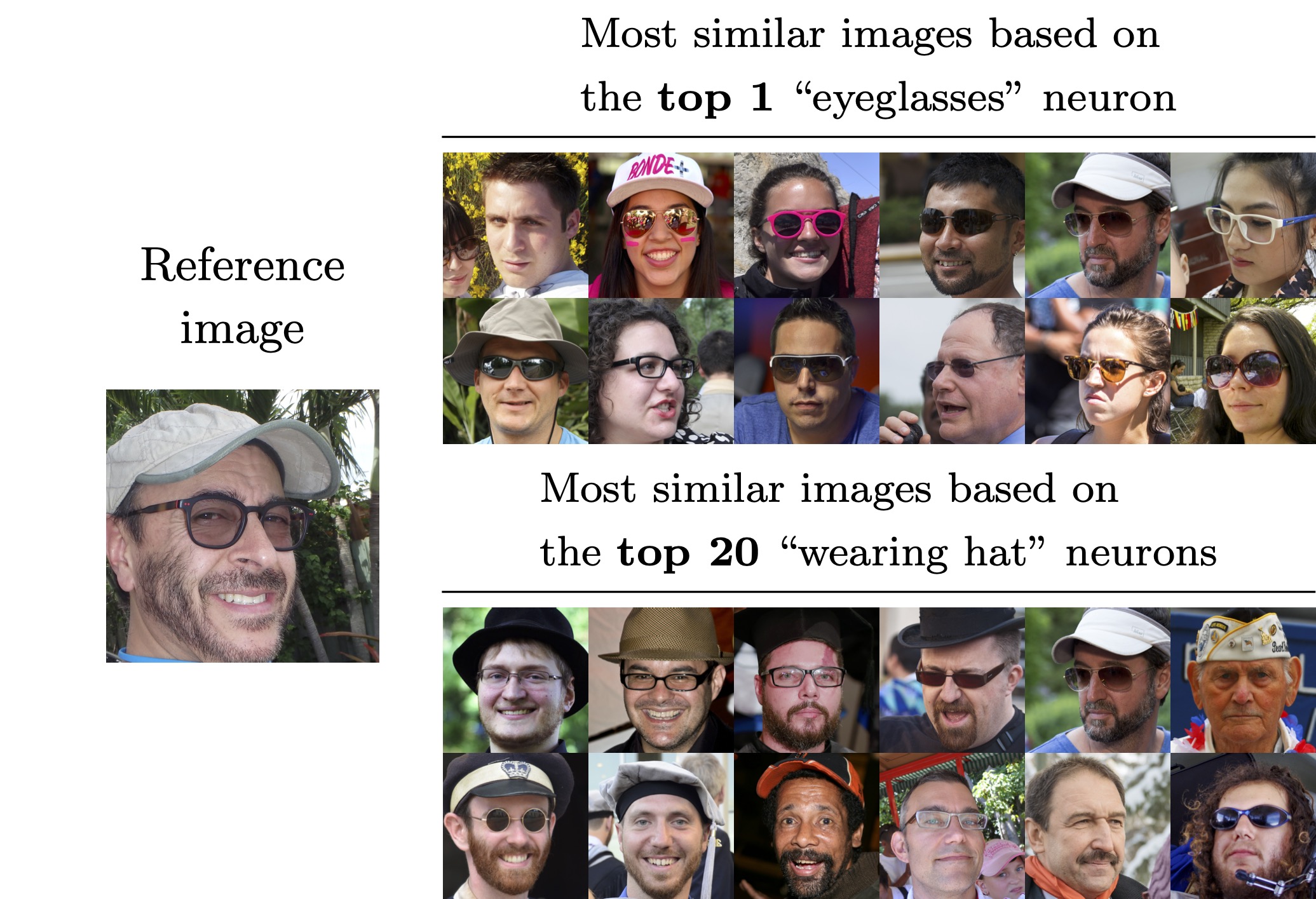}
	\caption{We seek to find the closest match to a reference image, on the left, from FFHQ, according to specific attributes -- here, \texttt{wearing hat} and \texttt{eyeglasses}.
		To do so, we select the neurons with the highest attribute sensitivity from CelebA and use them for comparisons in FFHQ, demonstrating the generalisability of these neurons.
		We show that very few neurons suffice to recover images with \texttt{eyeglasses} and \texttt{wearing hat}.}
	\label{fig:celeba_match}
	\vspace{-.5cm}
\end{figure}

\textbf{Finding similar images}
Qualitatively, we expect neurons to react to general and consistent features, which should also apply to comparing image pairs and different datasets.
To verify this, we compare image pairs from a different dataset, FFHQ~\cite{ffhq}, with and without specific attributes (\texttt{eyeglasses} and \texttt{wearing hat}) and select the neuron(s) with the highest \texttt{dna-emd} sensitivity on CelebA.
Using the selected neurons, we compare the \gls{dna-h} of a selected reference image to \glspl{dna-h} of 2000 random FFHQ samples.
We present our results in ~\cref{fig:celeba_match}.
We can verify that very few neurons are required to focus on high-level semantic attributes, even when selected on a different dataset.
We still observe some errors, possibly due to neurons reacting to several attributes simultaneously.

\begin{table*}
	\centering
	\resizebox{0.95\textwidth}{!}{
		\begin{tabular}{lccccccl}\toprule
			Feature extractor                                                 & \multicolumn{3}{c}{Mean target attribute $a$ sensitivity removal  $ \Delta_a$  (\%) $\nearrow$
			}                                                                 & \multicolumn{3}{c}{Mean other attributes sensitivity deviation $\frac{1}{|A| - 1} \sum_{b\in{}A\setminus{}a}| \Delta_b |$(\%) $\searrow$
			}
			\\\cmidrule(lr){0-0}\cmidrule(lr){2-4}\cmidrule(lr){5-7}
			                                                                  & Fr\'echet Distance                                                                                & DNA-Fr\'echet Distance & DNA-EMD & Fr\'echet Distance & DNA-Fr\'echet Distance & DNA-EMD \\
			Inception-v3~\cite{szegedyRethinkingInceptionArchitecture2016}     & 9.6                                                                                               & 94.8                 & 92.3           & 9.1                & 11.9                 & 10.9           \\
			CLIP image encoder (ViT-B/16)~\cite{pmlr-v139-radford21a}         & 20.1                                                                                              & 93.7                 & 94.3           & 17.6               & 7.2                  & 7.4            \\
			Stable Diffusion v1.4 encoder~\cite{Rombach_2022_CVPR}            & -                                                                                                 & 87.7                 & 81.4           & -                  & 19.4                 & 19.3           \\
			Random weights (ResNet-50)~\cite{ramanujanWhatHiddenRandomly2020} & 11.6                                                                                              & 72.1                 & 83.4           & 10.1               & 33.0                 & 20.1           \\
			DINO (ResNet-50)~\cite{caron2021emerging}                         & 15.8                                                                                              & 87.3                 & 93.5           & 8.9                & 16.2                 & 9.4            \\
			DINO (ViT-B/16)~\cite{caron2021emerging}                          & 19.0                                                                                              & 93.9                 & 94.2           & 16.3               & 10.2                 & 9.6            \\
			Mugs (ViT-B/16)~\cite{zhouMugsMultiGranularSelfSupervised2022}    & 20.0                                                                                              & 93.7                 & 95.5           & 16.7               & 10.3                 & 9.6            \\
			Mugs (ViT-L/16)~\cite{zhouMugsMultiGranularSelfSupervised2022}    & 34.6                                                                                              & 93.3                 & 95.3           & 28.0               & 9.4                  & 9.1            \\\midrule
			Mean                                                              & 18.7                                                                                              & 89.6                 & \textbf{91.2}  & 15.2               & 14.7                 & \textbf{11.9}  \\\bottomrule
		\end{tabular}
	}
	\caption{Customising different dataset comparison techniques to be insensitive to specific attributes. For each of the 40 attributes in CelebA, we use a weighted combination of distances over different layers or neurons with weights optimised such that the resulting distance between images with and without the attribute becomes zero. This is captured in the ``target attribute sensitivity removal'', which measures the relative drop in distance. We must ensure that the distance remains sensitive to the other 39 attributes. The ``other attributes sensitivity deviation'' measures the relative deviation in the original distance caused by the customisation. We show averages over all attributes on the CelebA testing set. \texttt{fd} can only combine distances over individual layers, making it challenging to ignore some attributes while preserving others. On the other hand, neuron-wise metrics such as \texttt{dna-fd} and \texttt{dna-emd} provide sufficient granularity for customising the distance to ignore one attribute while preserving the others. We only consider the latent space of Stable Diffusion v1.4, which we treat as a single layer -- hence we cannot perform a weighted combination of layers for the \texttt{fd} approach.}
	\label{tab:celeba-atts}
	\vspace{-.5cm}
\end{table*}

\subsection{Synthetic Data}
\label{subsec:stylegan}

Related systems have been used in the evaluation of synthetic image-creation techniques.
We thus qualitatively investigate the use of \texttt{dna-emd} to evaluate the quality -- \ie closeness to the distribution of real images -- of StyleGANv2~\cite{karrasAnalyzingImprovingImage2020a} generated images.
Here, we collect the \glspl{dna-h} for the datasets of real and generated images containing various classes~\cite{karrasAnalyzingImprovingImage2020a,yuLSUNConstructionLargescale2016,ffhq}.
We use these to select the most sensitive neurons (as above in~\cref{fig:celeba_match}) to differences between \texttt{real} and \texttt{fake} images, which we expect to be good indicators of realism.
These neurons are used to compare a separate dataset with generated images of one class not included in the datasets responsible for neuron selection -- \eg~when evaluating realism for cars, we select neurons based on cats, horses, churches, and faces, focusing on general realism rather than car-specific features.

\begin{figure*}[ht]
	\centering
	\includegraphics[width=1.0\linewidth, trim=0 0 0 0]{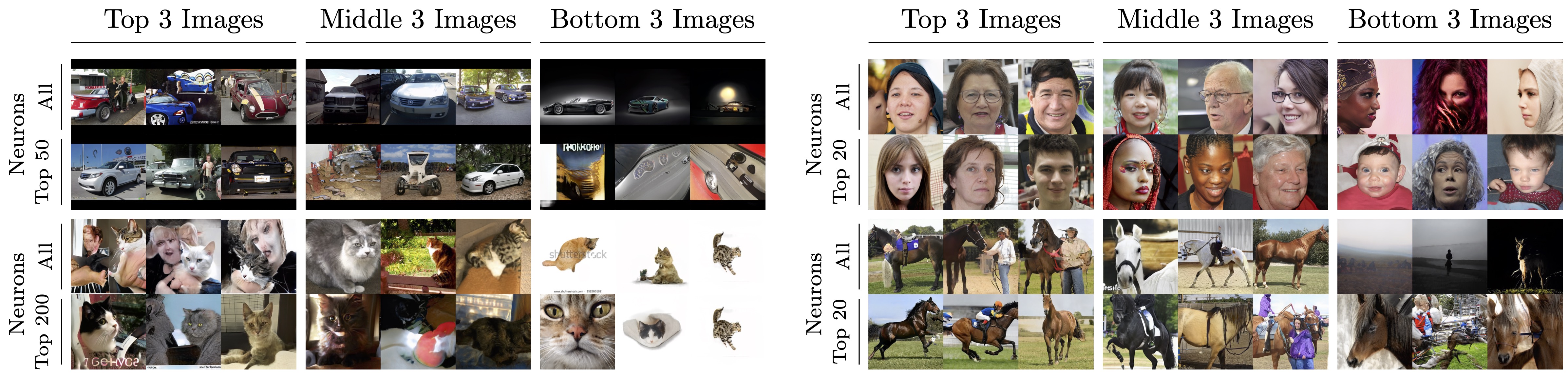}
	\caption{Generated StyleGANv2~\cite{karrasAnalyzingImprovingImage2020a} images of cars, faces, cats, and horses, ranked by \texttt{dna-emd} similarity to the corresponding real dataset's \gls{dna-h}.
		We selected neurons for realism by comparing real and synthetic images only featuring other classes (for general realism rather than focusing on class-specific differences) and
		compared this to using all neurons.
		Generally, selecting neurons results in rankings that better align with perceptual quality.}
	\label{fig:stylegan}
	\vspace{-.5cm}
\end{figure*}

Our results are reported in \cref{fig:stylegan}.
We clearly identify outliers in the generated samples using either selected or all neurons.
However, when using all neurons, top matches do not always match our perceptual quality assessment.
By selecting a small number of neurons reacting to realism, we favour images with fewer synthetic generation artefacts.

\subsection{Generalisation prediction under domain shifts}
Above, we have compared images and datasets from similar domains.
However, many applications require comparing datasets from distinct domains.
Here, we show the power of \glspl{dna} in cross-dataset generalisation prediction, which can serve, for instance, to select the best dataset for training when performing transfer learning.

For this, we compare our ranking of distances from dataset \glspl{dna} to the measured cross-dataset generalisation of a semantic-segmentation network reported in Tab. 3 of Lambert \etal~\cite{lambertMSegCompositeDataset2020}.
This reference provides mIoUs from an HRNet-W48~\cite{WangSCJDZLMTWLX19} semantic segmentation model architecture trained on seven datasets: ADE20K~\cite{zhouSceneParsingADE20K2017}, COCO~\cite{linMicrosoftCOCOCommon2014a}, BDD100K\cite{yuBDD100KDiverseDriving2020}, Cityscapes\cite{cordtsCityscapesDatasetSemantic2016b}, IDD~\cite{varmaIDDDatasetExploring2019}, Mapillary~\cite{neuholdMapillaryVistasDataset2017a}, and SUN-RGBD~\cite{songSUNRGBDRGBD2015}, and evaluated on all seven corresponding validation sets. Cross-generalization varies widely for different pairs of datasets, with mIoUs ranging from 0.2 (training on Cityscapes and validating on SUN-RGBD) to 69.7 (training on Mapillary and validating on Cityscapes).
Therefore, for each validation set $v \in V$, we have a ranking of which dataset's training sets transferred best in terms of mIoU.
We denote by $T^\text{gt}_v[i]$ the training set used by the model producing the $i$-th highest mIoU for validation set $v$, and by $\text{mIoU}_v(t)$ the mIoU observed with a model trained on $t$ and evaluated on the validation set $v$.

A good dataset distance metric will produce similar mIoU rankings -- and importantly, without training a model.
We therefore compare all pairs of datasets using \texttt{fd}, \texttt{dna-fd}, and \texttt{dna-emd}, and rank them by distance.
We denote by $T_v^\text{pred}[i]$ the training set ranked at the $i$-th position when compared to validation set $v$.
To aggregate results, we measure the discrepancy $d$ between predicted and reference rankings using average mIoU differences:
\begingroup
\setlength{\abovedisplayskip}{5pt}
\setlength{\belowdisplayskip}{5pt}
\begin{equation}
	d = \frac{1}{|V||T|} \sum_{v \in V} \sum_{i=1}^{|T^\text{gt}_v|} | \text{\footnotesize mIoU}_v(T_v^\text{gt}[i]) - \text{\footnotesize mIoU}_v(T_v^\text{pred}[i])|
\end{equation}
\endgroup
This discrepancy penalises out-of-rank predictions based on the difference of mIoU at those ranks.

In addition to the Mugs feature extractor, we also consider domain-specific feature extractors. We evaluate cross-dataset generalisation using features extracted from an HRNet-W48 semantic segmentation model trained on MSeg~\cite{lambertMSegCompositeDataset2020} which combines all datasets used in the experiment. We also use HRNet-W48 models trained on the validation domains. We report results relying on the features from the last layer of each model.
We present the summary results for different feature extractors and metrics in \Cref{tab:mseg-summary}.

Using \texttt{dna-emd} with a self-supervised network provides the best cross-dataset generalization.
While being specifically adapted to the task and datasets considered, HRNet-W48 models fail to perform as well, likely due to the less general features not allowing to measure domain shifts as well.
The average mIoU error in ranking datasets with \texttt{dna-emd} with Mugs features is only 0.76, indicating very good predictions of cross-generalization performance without training a model, markedly superior to \texttt{fd} and \texttt{dna-fd}.

\begin{table}[h]
	\centering
	\resizebox{1.0\columnwidth}{!}{
		\begin{tabular}{lccc}\toprule
			                   Feature extractor        & Fr\'echet Distance                 & DNA-Fr\'echet Distance       & DNA-EMD      \\\cmidrule(lr){0-0} \cmidrule(lr){2-4}
			Mugs (ViT-B/16)            & 1.66                 & 1.79      & \textbf{0.76}               \\
			HRNet-W48 (all domains)             & 9.63                 & 11.18     & 9.40     \\
			HRNet-W48 (val. domain)           & 13.9                 & 14.5      & 6.85      \\\cmidrule(lr){1-4}
			Random ordering            &   \multicolumn{3}{c}{ 14.93 $\pm$ 1.86  (50 samples)      }  
			\\\bottomrule
		\end{tabular}
	}
	\caption{Effectiveness of using dataset comparisons to predict semantic segmentation transfer learning performance. We compare the ranking of training datasets by a model's transfer learning performance to the ranking of datasets based on their distance to the validation set. We measure the severity of errors in predicted ranking by calculating the average difference in mIoU scores of reference models when ranked by mIoU and when ranked by the distance between their training and validation sets.}

	\label{tab:mseg-summary}
	\vspace{-.65cm}
\end{table}


\section{Limitations}
\textbf{Labelled data requirements for neuron selection} Our neuron selection experiments in this work rely on labelled images to find neuron combination strategies. This is not always available, in which case unsupervised clustering techniques such as deepPIC~\cite{jaipuriaDeepPICDeepPerceptual2022} could be used.

\textbf{Combining neurons}
Many neurons are likely to be polysemantic~\cite{nelsonelhageToyModelsSuperposition2022,olahZoomIntroductionCircuits2020}, meaning that they are likely to react to multiple unrelated inputs. The approaches used in this paper to combine information from different neurons might be too limited to properly isolate specific attributes.

\textbf{Discarded information in the DNA representation} To make \glspl{dna} practical and scalable, we have discarded information about features. This includes spatial information about where activations occur and dependencies between activations of all neurons. These could help to obtain an even better representation.

\section{Conclusion}
We have presented a general and granular representation for images. This representation is based on keeping track of distributions of neuron activations of a pre-trained feature extractor. One DNA can be created from a single image or a complete dataset.
Image DNAs are compact and granular representations which require no training, hyperparameter tuning, or labelling, regardless of the type of images considered.
Our experiments have demonstrated that even with simplistic comparison strategies, DNAs can provide valuable insights into attribute-based comparisons, synthetic image quality assessment, and dataset differences.

\blfootnote{\hspace{-2em} \textbf{Acknowledgements}
	This work was supported by EPSRC Programme Grant ``From Sensing to Collaboration'' (EP/V000748/1), the EPSRC Centre for Doctoral Training in Autonomous Intelligent Machines and Systems [EP/S024050/1], and Oxbotica.
	The authors would like to acknowledge the use of Hartree Centre resources and the University of Oxford Advanced Research Computing (ARC) facility in carrying out this work.
	We thank Adam Caccavale, Pierre-Yves Lajoie, Pierre Osselin, and David Williams for helpful discussions and inputs that contributed to this work.
}
{\small
	\bibliographystyle{ieee_fullname}
	\bibliography{dna-bib}
}
\clearpage
\appendix
\onecolumn
\section{Technical details}
\subsection{Feature extractors}
\subsubsection{Features considered}

We present details of the locations of neurons considered in \cref{tab:neurons-used}. For the Stable diffusion v1.4 encoder, we expect the latent space to contain sufficient information to reconstruct the original image. As such, we only consider the latent space, treating each element as a neuron.

\begin{table}[htb]
	\centering
	\begin{tabular}{@{}lcc@{}}
		\toprule
		Feature extractor architecture & Layers considered                                                                                                                                                       & Number of neurons per layer                                                   \\ \midrule
		Inception-v3                   & \multicolumn{1}{p{6cm}}{Outputs of the first and second max pooling layers, input features to the auxiliary classifier,  and output of the final average pooling layer} & $64$ / $192$ / $768$ / $2048$                                                 \\\addlinespace[0.5em]
		Stable diffusion v1.4 encoder  & \multicolumn{1}{p{6.5cm}}{Latent space produced by the VAE encoder}                                                                                                     & $4096$                                                                        \\\addlinespace[0.5em]
		ResNet-50                      & \multicolumn{1}{p{6.5cm}}{Output of each block}                                                                                                                         & \multicolumn{1}{p{5cm}}{\centering  $3 \times [256]$ / $4 \times [512]$ /     \\ $6 \times [1024]$ /  $3 \times [2048]$   }          \\\addlinespace[0.5em]
		Vision Transformer (ViT-B/16)  & \multicolumn{1}{p{6.5cm}}{Output of each of the 12 self-attention layers, and output of the final normalisation layer}                                                  & $13 \times [768]$                                                             \\\addlinespace[0.5em]
		Vision Transformer (ViT-L/16)  & \multicolumn{1}{p{6.5cm}}{Output of each of the 24 self-attention layers, and output of the final normalisation layer}                                                  & $25 \times [1024]$                                                            \\ \addlinespace[0.5em]
		HRNet-W48                      & \multicolumn{1}{p{6.5cm}}{Outputs of each stage, where we treat each branch of the stage as providing a different layer. We also include the output of the classifier}  & \multicolumn{1}{p{5cm}}{\centering $256$ / $(48 / 96)$ / $(48 / 96 / 192)$  / \\ $(48 / 96 / 192 / 384)$ / $194$   }                    \\ \bottomrule
	\end{tabular}
	\caption{Details of layers and neurons considered for all architectures.}
	\label{tab:neurons-used}
\end{table}

\subsubsection{Spatial accumulation for Vision Transformers}
In \cref{sec:dna-math}, we describe feature maps as having spatial dimensions $S_h^l \times S_w^l$. In the case of vision transformers, we treat the patch/token dimensions as the only spatial dimension. Activations over the patch/token dimension are accumulated in the histogram of the corresponding neuron.

\subsection{Data processing}
\paragraph{Activation normalisation}
As discussed in \cref{sec:settings}, activations at each neuron are normalised to ensure comparable scales over different neurons. The normalisation relies on tracking minimum and maximum activation values over the following datasets: StyleGANv2 generated images of cars, cats, churches, faces, and horses, LSUN cars, LSUN cats, LSUN churches, LSUN horses, FFHQ,  Metfaces, CelebA, Cityscapes, KITTI, IDD, ADE20K, BDD100K, Mapillary, Wilddash, COCO, and SUN-RGBD.
Using the minimum and maximum values observed, $\mathtt{a}_\text{min}$ and $\mathtt{a}_\text{max}$, for each neuron over all of these datasets, we recompute normalised activations $\mathtt{a}_\text{norm}$ from the activations observed during inference $\mathtt{a}$ to produce histograms as follows:
\begin{align}
	 & \mu = \frac{1}{2}(\mathtt{a}_\text{max} + \mathtt{a}_\text{min})     \\
	 & \sigma = \frac{1}{2}(\mathtt{a}_\text{max} - \mathtt{a}_\text{min} ) \\
	 & \mathtt{a}_\text{norm} = \frac{\mathtt{a} - \mu }{\sigma}
\end{align}

\paragraph{Cropping}
Images are centre-cropped to squares and resized depending on the feature extractor, as described in \cref{tab:crop-sizes}.

\begin{table}[htb]
	\centering
	\begin{tabular}{@{}lc@{}}
		\toprule
		Feature extractor             & Image size used    \\ \midrule
		Inception-v3                  & $ 299 \times 299 $ \\
		CLIP image encoder (ViT-B/16) & $ 224 \times 224 $ \\
		Stable diffusion v1.4 encoder & $ 256 \times 256 $ \\
		Random weights (ResNet-50)    & $ 224 \times 224 $ \\
		DINO (ResNet-50)              & $ 224 \times 224 $ \\
		DINO (ViT-B/16)               & $ 224 \times 224 $ \\
		Mugs (ViT-B/16)               & $ 224 \times 224 $ \\
		Mugs (ViT-L/16)               & $ 224 \times 224 $ \\
		HRNet-W48                     & $ 360 \times 360 $ \\\bottomrule
	\end{tabular}
	\caption{Image sizes used for different feature extractors.}
	\label{tab:crop-sizes}
\end{table}

\subsection{Random image transformations for Cityscapes retrieval task}
In \cref{subsec:cs-retrieval}, we attempt to retrieve randomly augmented Cityscapes images. The augmentations applied to each image are 2 randomly selected operations applied sequentially. The operations considered are \texttt{Identity}, \texttt{Shear X},  \texttt{Shear Y},  \texttt{Translate X},  \texttt{Translate Y}, \texttt{Rotate}, \texttt{Adjust brightness}, \texttt{Adjust saturation}, \texttt{Adjust contrast}, \texttt{Adjust sharpness}, \texttt{Posterize}, \texttt{Auto contrast}, \texttt{Equalize}, \texttt{Salt-and-pepper noise}, \texttt{Gaussian noise}, and \texttt{Blur}.

\subsection{Weights optimisation for attribute removal}
For the optimisation of weights used in \cref{sec:celeba-neuron-selection}, we use the Adam optimiser with a learning rate of
$1\mathrm{e}{-5}$, and first- and second-moment decays rates
$\beta_1 = 0.9$ and $\beta_2 = 0.999$. Weights are initialised to a constant value of $\frac{1}{\text{number of neurons}}$.

We perform optimisation of the loss in \cref{eq:loss} on DNAs from the training set, which were generated using up to $50000$ images for each DNA. We use early stopping by monitoring the loss on the validation set, stopping after $50$ iterations without improvements. The results in \cref{tab:celeba-atts} are then reported on the testing set.

\subsection{Neuron selection for StyleGANv2 synthetic images}
In \cref{subsec:stylegan}, we select neurons to rank images from one class by selecting the most sensitive neurons to differences between \texttt{real} and \texttt{fake} images of all other classes.
More precisely, this is done by creating a \gls{dna-h} representing all \texttt{real} images from the other classes, and a DNA representing all \texttt{fake} images from the other classes. Combining the \gls{dna-h} of multiple datasets is done by summing the counts for each histogram from the \glspl{dna-h} to combine. This is equivalent to creating a new single \gls{dna-h} using all the images from the different datasets.
The sensitivity of neurons is then computed using the neuron-wise \gls{emd} between the combined \texttt{real} \gls{dna-h} and the combined \texttt{fake} \gls{dna-h}.

\subsection{Example histograms}
\begin{figure}[htb]
	\centering
	\begin{subfigure}[b]{0.6\linewidth}
		\centering
		\includegraphics[width=0.49\linewidth]{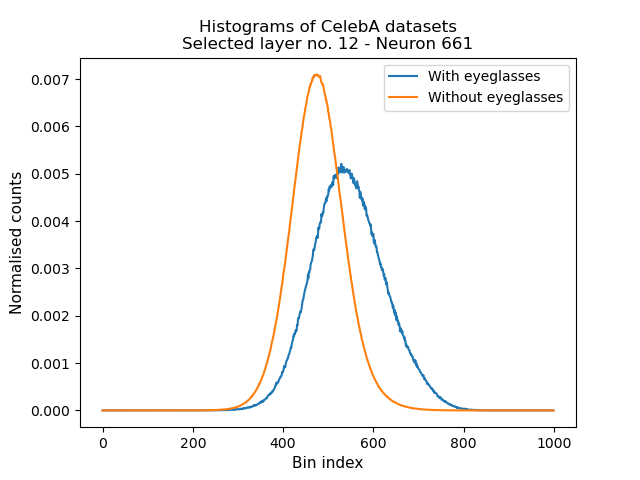}
		\includegraphics[width=0.49\linewidth]{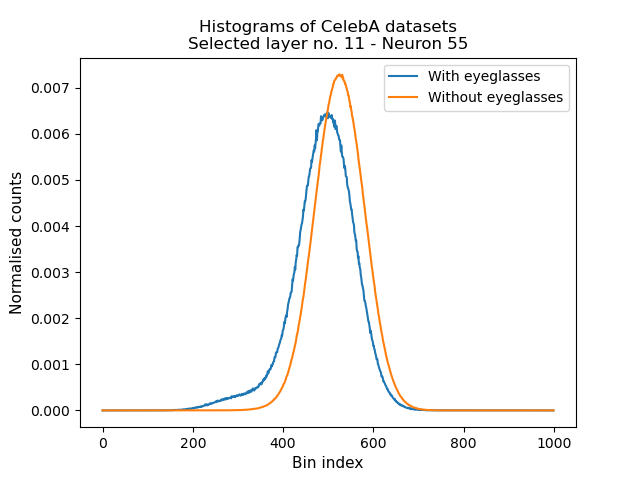}
		\caption{Examples of histograms resulting in the largest Earth Mover's Distances }
	\end{subfigure}
	\\
	\begin{subfigure}[b]{0.6\linewidth}
		\centering
		\includegraphics[width=0.49\linewidth]{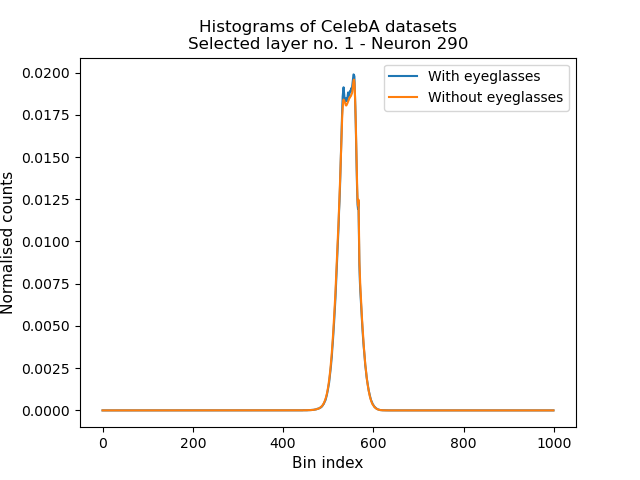}
		\includegraphics[width=0.49\linewidth]{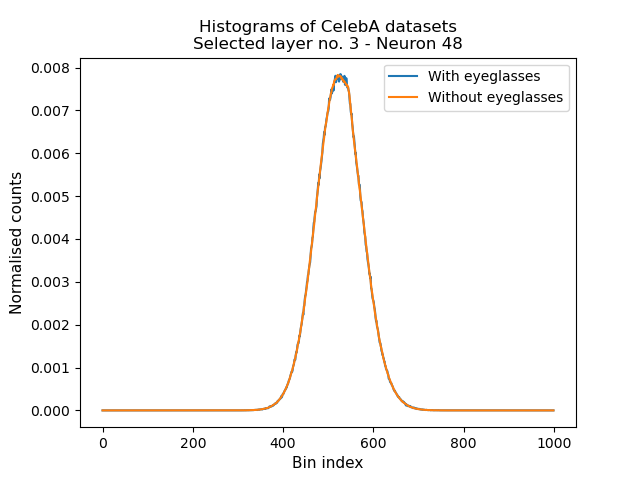}
		\caption{Examples of histograms resulting in the lowest Earth Mover's Distances}
	\end{subfigure}
	\caption{Visualisation of normalised histograms for specific neurons from \glspl{dna-h} (Mugs ViT-B/16) of images from the CelebA dataset with and without \textbf{\texttt{eyeglasses}}.}
	\label{fig:hists-plots-eyeglasses}
\end{figure}

\begin{figure}[htb]
	\centering
	\begin{subfigure}[b]{0.6\linewidth}
		\centering
		\includegraphics[width=0.49\linewidth]{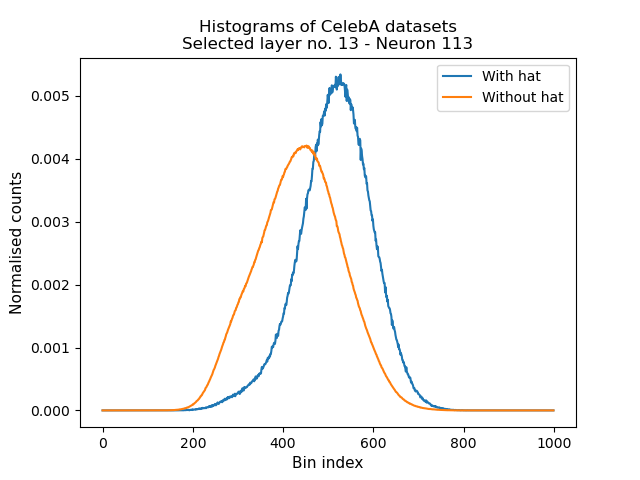}
		\includegraphics[width=0.49\linewidth]{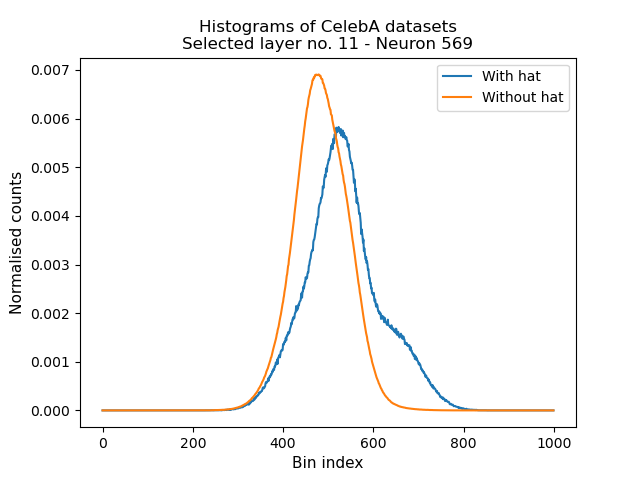}
		\caption{Examples of histograms resulting in the largest Earth Mover's Distances }
	\end{subfigure}
	\\
	\begin{subfigure}[b]{0.6\linewidth}
		\centering
		\includegraphics[width=0.49\linewidth]{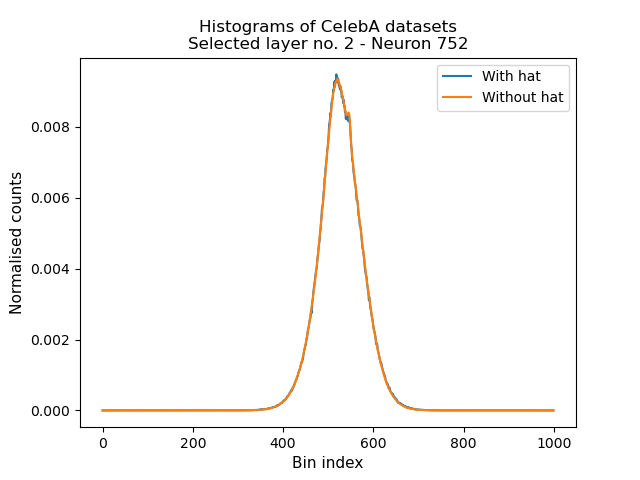}
		\includegraphics[width=0.49\linewidth]{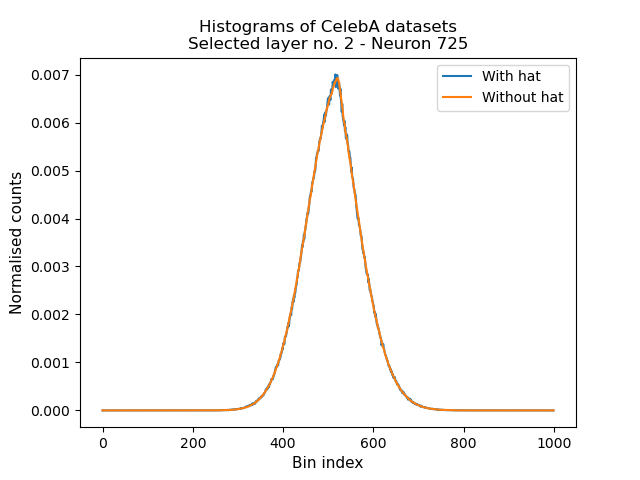}
		\caption{Examples of histograms resulting in the lowest Earth Mover's Distances}
	\end{subfigure}
	\caption{Visualisation of normalised histograms for specific neurons from \glspl{dna-h} (Mugs ViT-B/16) of images from the CelebA dataset with and without \textbf{\texttt{wearing hat}}.}
	\label{fig:hists-plots-hats}
\end{figure}

In \cref{fig:hists-plots-eyeglasses,fig:hists-plots-hats}, we present examples of histograms from specific neurons of the \gls{dna-h} of CelebA images with or without some attributes.
We observe that their shapes would not always be well approximated by a Gaussian distribution, as is done with \glspl{dna-g}.

\section{Additional results details}
The following sections present more detailed results. Unless mentioned otherwise, the results use \texttt{dna-emd} with a Mugs (ViT-B/16) feature extractor.

\subsection{Comparing images to a reference dataset with different neurons}
\begin{figure}[htb]
	\centering
	\begin{subfigure}[b]{0.9\linewidth}
		\centering
		\includegraphics[width=\linewidth]{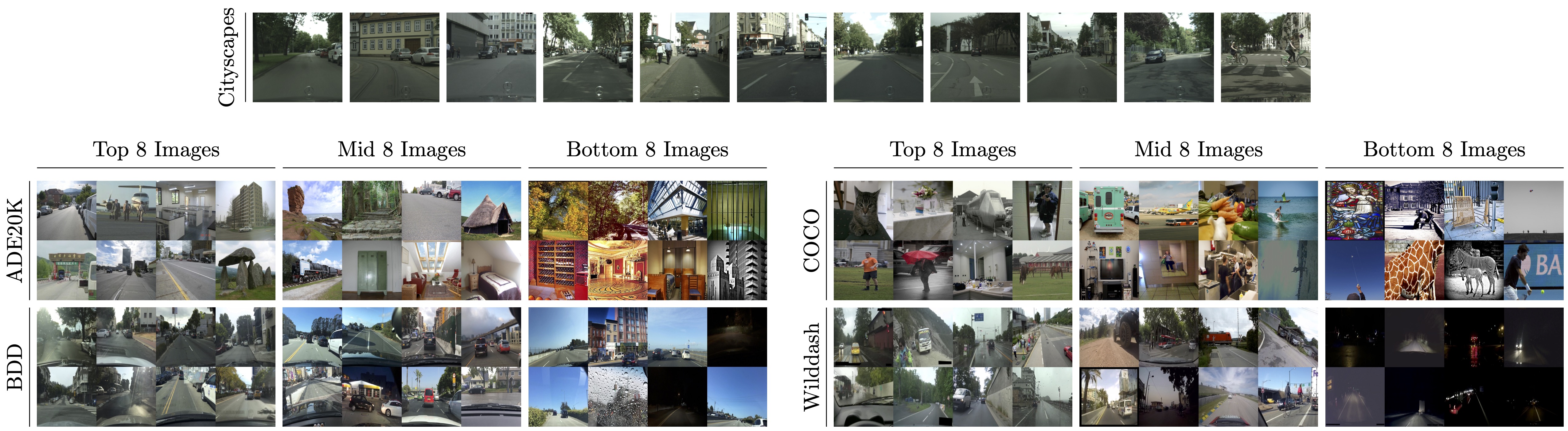}
		\caption{Using all neurons from the \textbf{first layer} of the Mugs (ViT-B/16) feature extractor.}
	\end{subfigure}
	\\
	\begin{subfigure}[b]{0.9\linewidth}
		\centering
		\includegraphics[trim={0 0 0 15cm},clip, width=\linewidth]{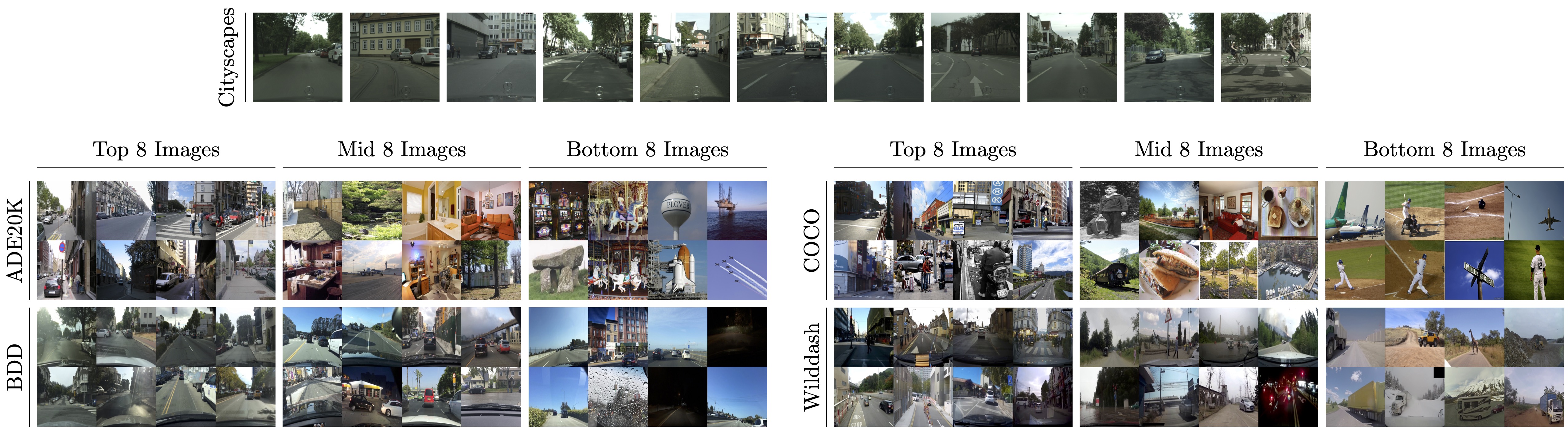}
		\caption{Using all neurons from the \textbf{last layer} of the Mugs (ViT-B/16) feature extractor.}
	\end{subfigure}
	\caption{Images from different datasets organised by \texttt{dna-emd} when compared to Cityscapes~\cite{cordtsCityscapesDatasetSemantic2016b} using different neurons from the feature extractor.}
	\label{fig:supp-mseg}
\end{figure}

In \cref{fig:supp-mseg}, we present ranked images from different datasets with specific neurons from their \glspl{dna-h} compared to the \gls{dna-h} of the Cityscapes dataset. Compared to \cref{fig:cs_i2d} in which all neurons are considered, here we show results with neurons from the first and last selected layers of the feature extractor.
When using neurons from the first layer, colours and textures appear much more important in producing the score. The best matches sometimes do not correspond to similar types of scenes, such as in COCO, but display similar colour profiles. Worst matches tend to contain high-frequency patterns or few features.
On the other hand, when using neurons from the last layer, semantic content seems to be the main factor. Top-ranking images always show the same type of environment but do not always have the same colour profiles.

\subsection{CelebA attribute sensitivity removal}
\subsubsection{Visualisation of \texttt{dna-emd} neuron-wise differences and weights learned}

\begin{figure}[htb]
	\centering
	\begin{subfigure}[b]{0.7\linewidth}
		\centering
		\includegraphics[width=0.49\linewidth]{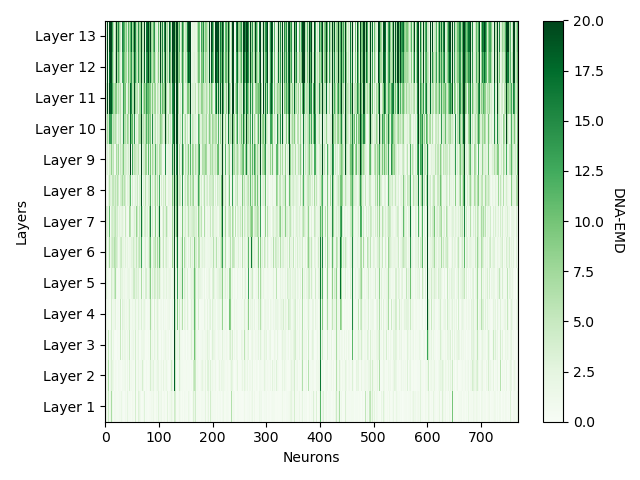}
		\includegraphics[width=0.49\linewidth]{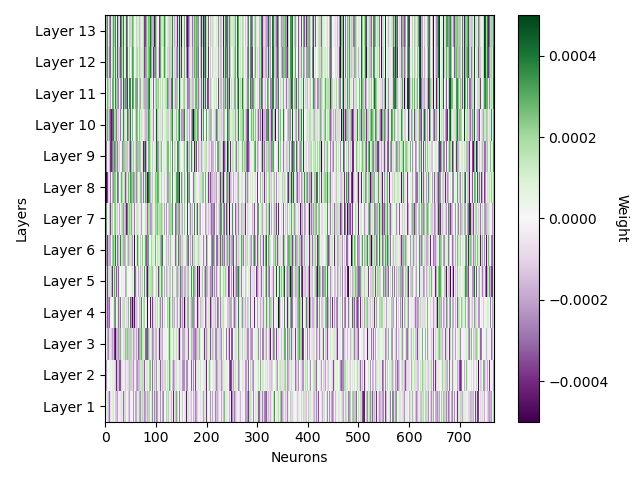}
		\caption{Distances (\textbf{left}) and optimised weights (\textbf{right}) for the \textit{eyeglasses} attribute.}
	\end{subfigure}
	\\
	\begin{subfigure}[b]{0.7\linewidth}
		\centering
		\includegraphics[width=0.49\linewidth]{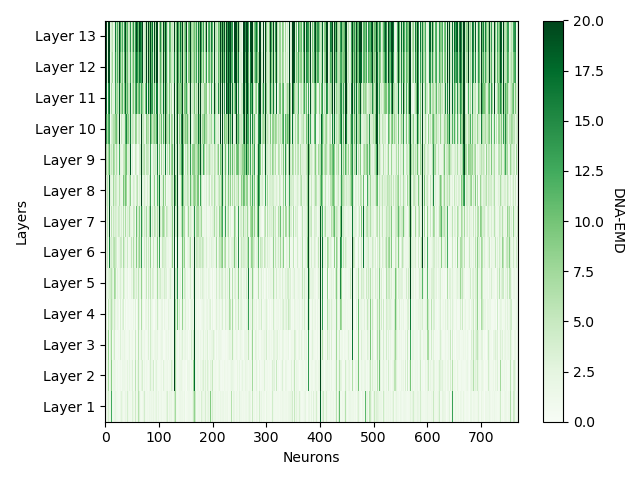}
		\includegraphics[width=0.49\linewidth]{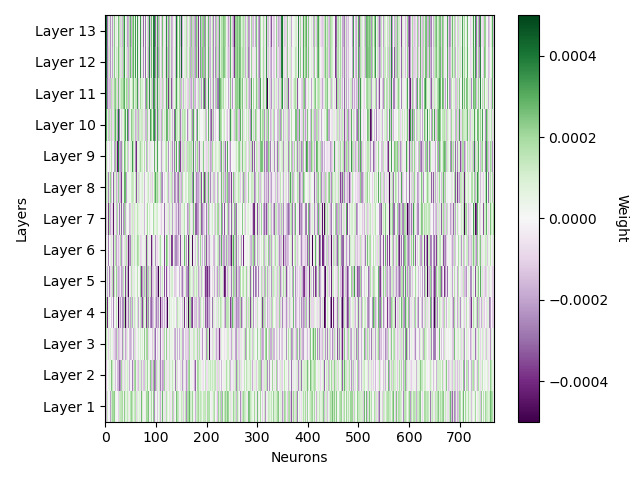}
		\caption{Distances (\textbf{left}) and optimised weights (\textbf{right}) for the \textit{wearing hat} attribute.}
	\end{subfigure}
	\caption{Comparison of \texttt{dna-emd} distances between \glspl{dna-h} with and without an attribute, and optimised weights for attribute removal.}
	\label{fig:heatmaps}
\end{figure}

In \cref{fig:heatmaps}, we compare the neuron-wise \gls{emd} between \glspl{dna-h} of datasets with and without specific attributes to the weights learned in \cref{sec:celeba-neuron-selection}.
Distances appear larger at later layers, but the attention from the weights allows us to focus on specific neurons spread over all layers and thus ignore the attribute, highlighting the need for granularity and a multi-layered approach.

\subsubsection{Standard deviations of scores}
\cref{tab:celeba-atts-stds} details the standard deviations computed over all forty attributes for the results in \cref{tab:celeba-atts} in~\cref{sec:celeba-neuron-selection}.
\begin{table}[ht]
	\centering
	\resizebox{0.95\textwidth}{!}{
		\begin{tabular}{lccccccl}\toprule
			Feature extractor                                                 & \multicolumn{3}{c}{Target attribute sensitivity removal $\Delta_\text{rem}$  std. dev.  (\%)
			}                                                                 & \multicolumn{3}{c}{Other attributes sensitivity deviation $|\Delta_\text{rem}|$ std. dev. (\%)
			}
			\\\cmidrule(lr){0-0}\cmidrule(lr){2-4}\cmidrule(lr){5-7}
			                                                                  & Fr\'echet Distance                                                                               & DNA-Fr\'echet Distance & DNA-EMD  & Fr\'echet Distance & DNA-Fr\'echet Distance & DNA-EMD \\
			Inception-v3~\cite{szegedyRethinkingInceptionArchitecture2016}     & 7.55                                                                                             & 4.30                 & 5.65           & 7.40               & 6.99                 & 6.37          \\
			CLIP image encoder (ViT-B/16)~\cite{pmlr-v139-radford21a}         & 13.84                                                                                            & 5.78                 & 4.90           & 12.80              & 4.24                 & 3.59          \\
			Stable Diffusion v1.4 encoder~\cite{Rombach_2022_CVPR}            & -                                                                                                & 9.62                 & 11.05          & -                  & 6.83                 & 6.32          \\
			Random weights (ResNet-50)~\cite{ramanujanWhatHiddenRandomly2020} & 12.13                                                                                            & 28.05                & 19.58          & 11.91              & 17.12                & 9.97          \\
			DINO (ResNet-50)~\cite{caron2021emerging}                         & 13.03                                                                                            & 13.28                & 5.19           & 6.91               & 9.72                 & 4.34          \\
			DINO (ViT-B/16)~\cite{caron2021emerging}                          & 12.26                                                                                            & 4.50                 & 4.19           & 12.47              & 5.62                 & 4.83          \\
			Mugs (ViT-B/16)~\cite{zhouMugsMultiGranularSelfSupervised2022}    & 12.18                                                                                            & 5.47                 & 4.76           & 11.62              & 6.33                 & 5.39          \\
			Mugs (ViT-L/16)~\cite{zhouMugsMultiGranularSelfSupervised2022}    & 17.89                                                                                            & 5.18                 & 4.28           & 16.91              & 5.71                 & 4.62          \\\bottomrule
		\end{tabular}
	}
	\caption{Standard deviations over all forty attributes for scores presented in \cref{tab:celeba-atts}.}
	\label{tab:celeba-atts-stds}
\end{table}

\subsubsection{Detailed deviations}
\begin{figure}[htb]
	\centering
	\includegraphics[width=0.8\linewidth]{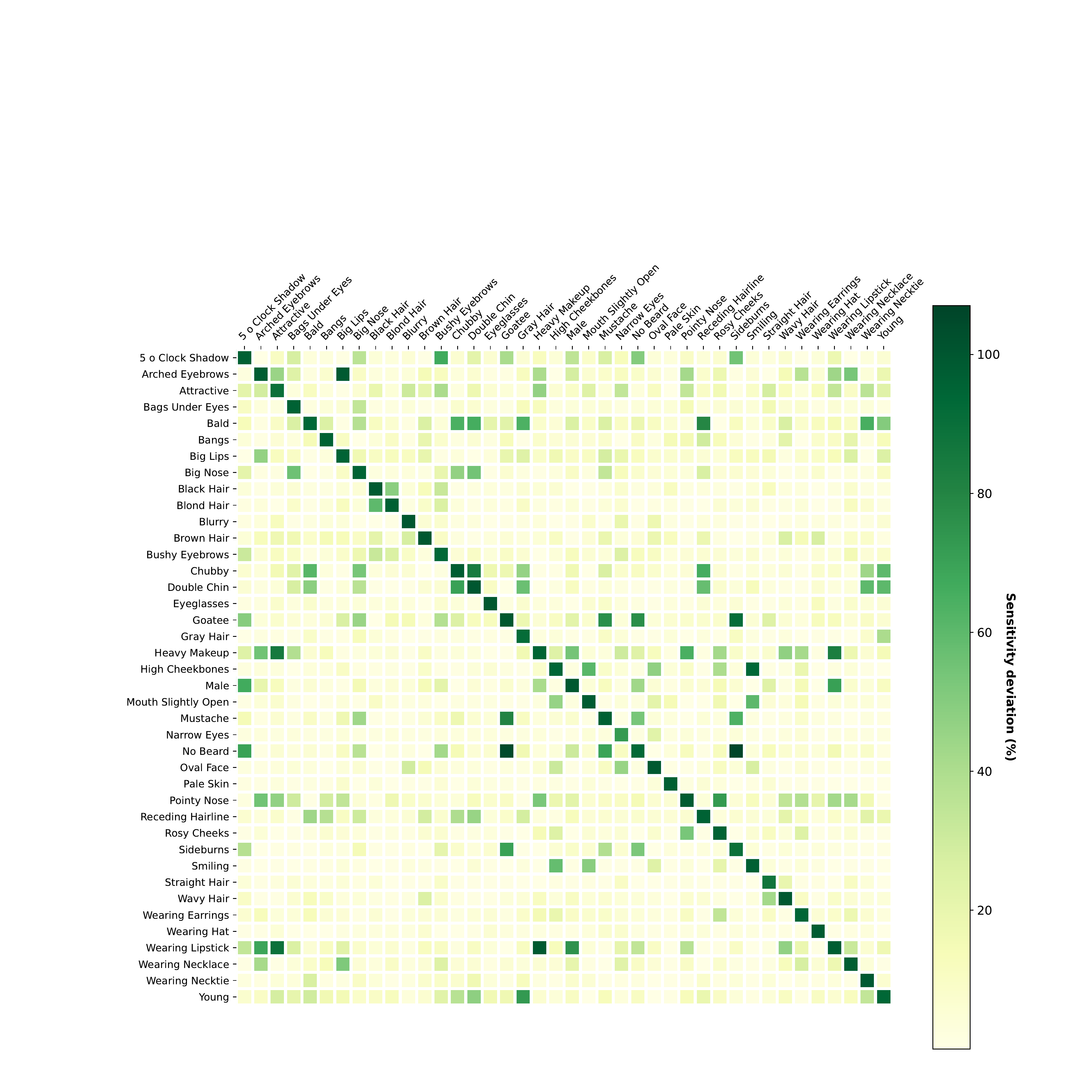}
	\caption{For each ignored attribute in the rows, we show the sensitivity deviation of all other attributes from the columns. The diagonals describe the relative drop in distances on the ignored attribute.}
	\label{fig:all-deviations}
\end{figure}

In \cref{sec:celeba-neuron-selection}, we weighted distances over different neurons to remove sensitivity to one attribute while preserving others. In \cref{fig:all-deviations}, we visualise which attributes deviate most when ignoring another.
We see that some attributes are particularly challenging to disentangle, often when we expect them to be correlated. For example, when ignoring the \texttt{no beard} attribute, we cause a large deviation in the \texttt{goatee} attribute. We might expect these to react to similar neurons.
Still, we believe improvements in the optimised loss might help reduce this entanglement.

\begin{figure}[htb]
	\centering
	\includegraphics[width=0.8\linewidth]{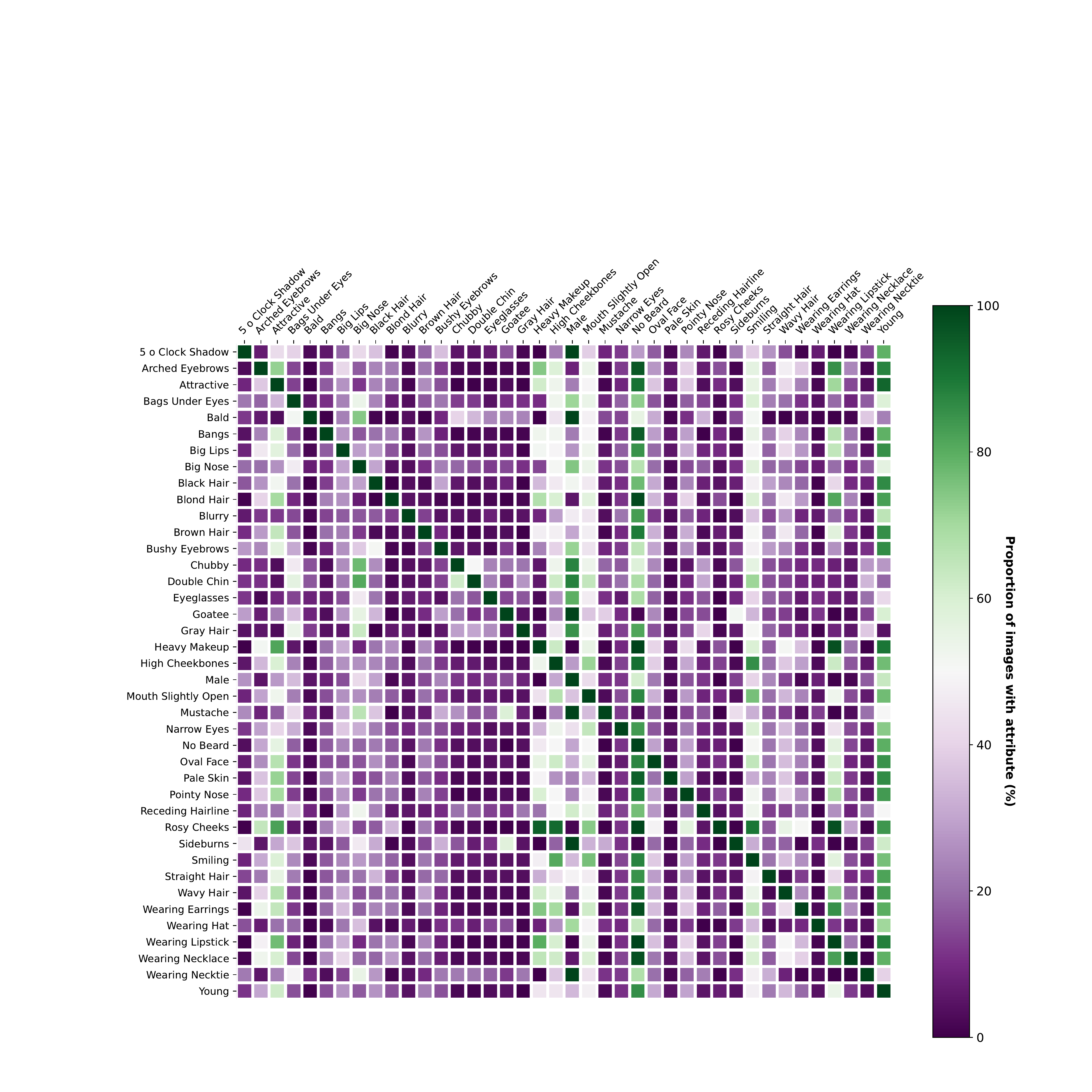}
	\caption{Illustration of correlations between attributes in the CelebA dataset. For each attribute in the rows, we show the percentage of images containing that attribute that also possess the attribute in the columns. Attributes can be negatively correlated when we observe values close to $0\%$, or positively correlated with values close to $100\%$.}
	\label{fig:celeba-stats}
\end{figure}

We also visualise overlaps between attributes in the CelebA dataset in \cref{fig:celeba-stats}.

\subsection{FFHQ image pair comparisons}
\begin{figure}[htb]
	\centering
	\includegraphics[width=0.9\linewidth]{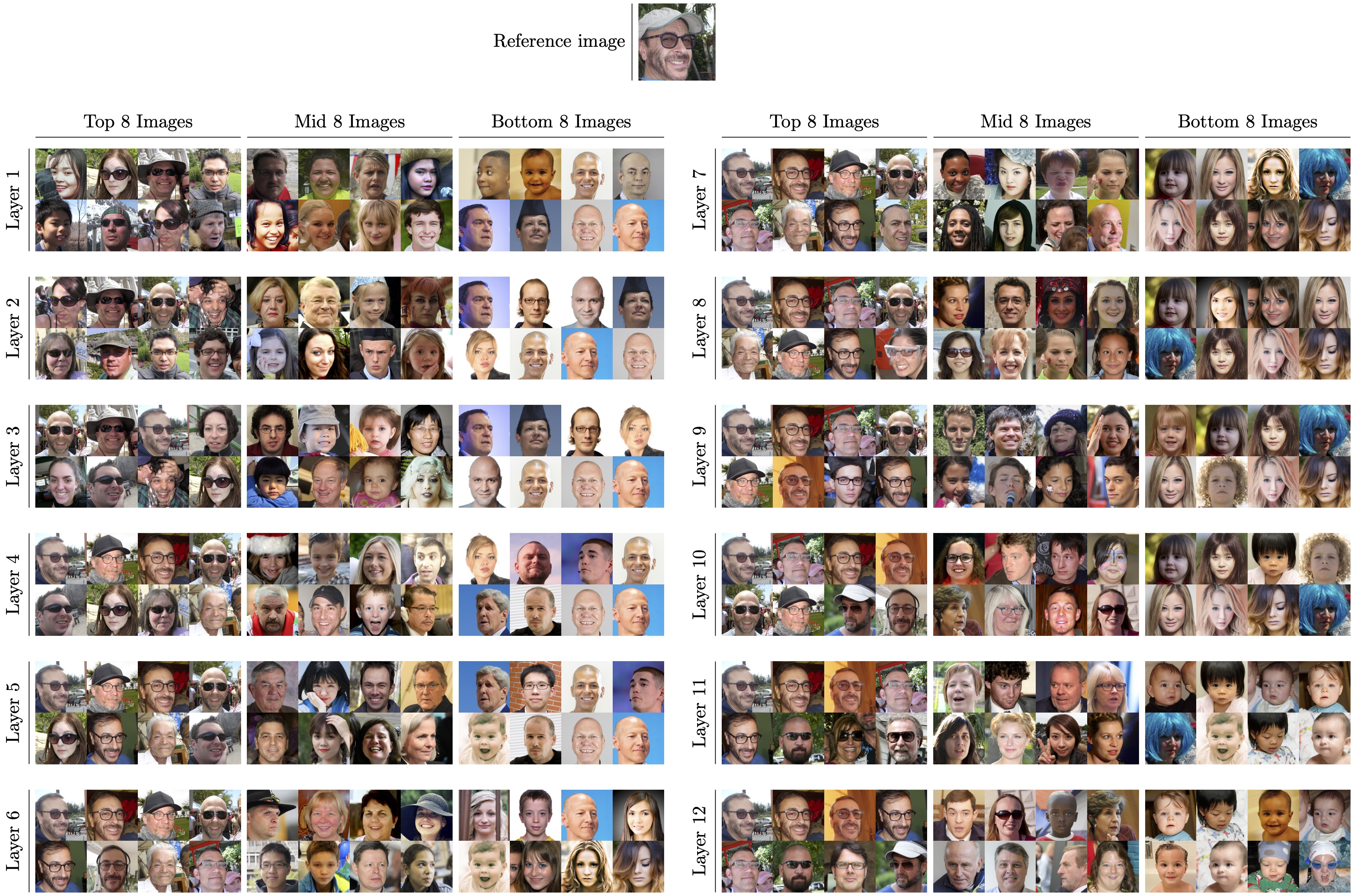}
	\caption{Ranked matches to the reference image using \texttt{dna-emd} with neurons from \textbf{different layers} of the Mugs (ViT-B/16) feature extractor.}
	\label{fig:supp-faces-layers}
\end{figure}

In this section, we present additional details for the results shown in \cref{fig:celeba_match}. In addition to showing middle and bottom-ranked matches, we also consider different neurons for comparisons.
\cref{fig:supp-faces-layers} presents the matches ranked using all neurons from different layers of the feature extractor.
Top matches from the first layer do not always focus on having similar semantic attributes. However, they all contain similar backgrounds and colours. Top matches from the last layer have much more diverse colour profiles and better match other images of the person in the reference image.

\begin{figure}[htb]
	\centering
	\includegraphics[width=0.9\linewidth]{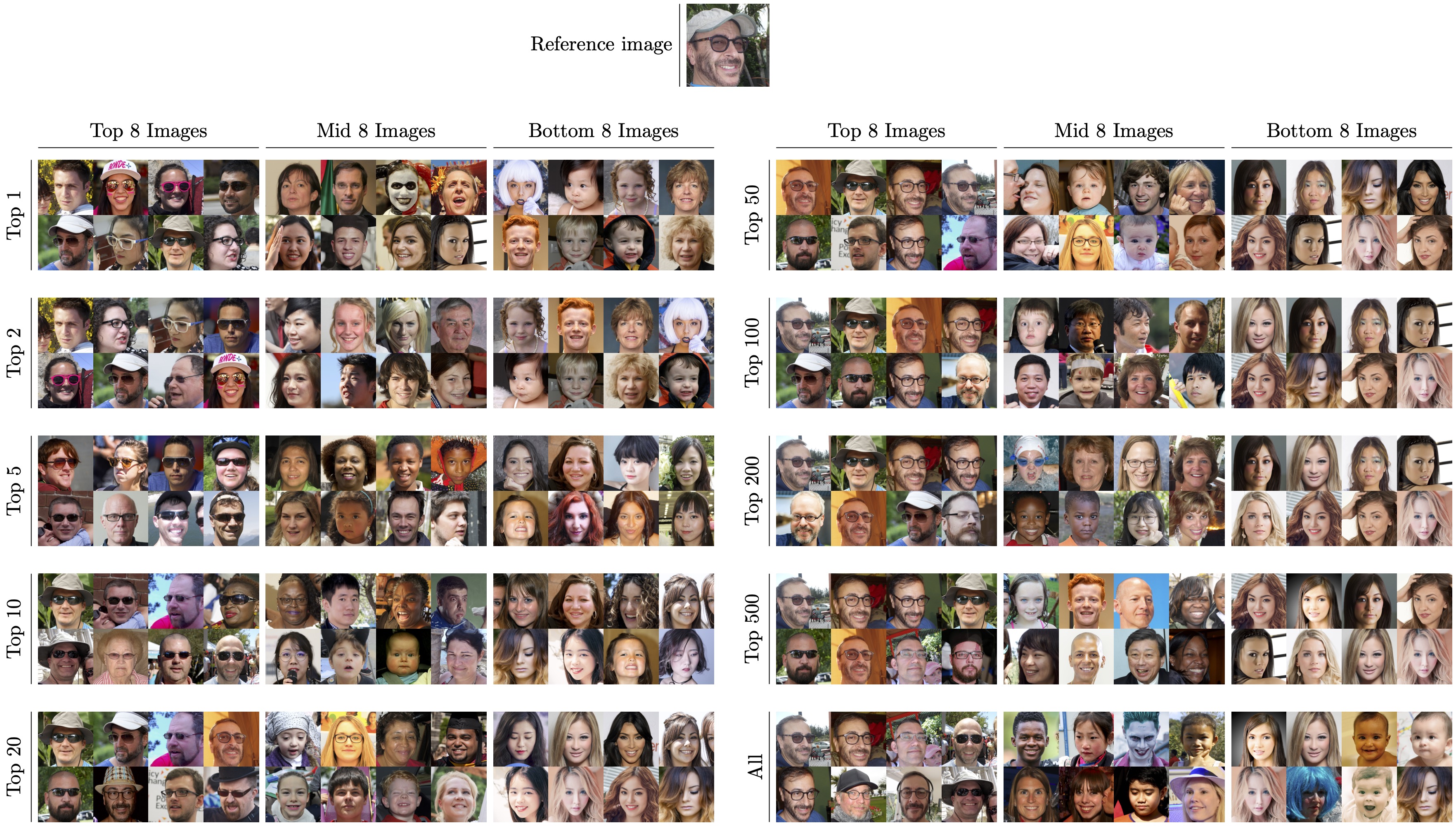}
	\caption{Ranked matches to the reference image using \texttt{dna-emd} with neurons of the Mugs (ViT-B/16) feature extractor sensitive to the \textbf{\texttt{eyeglasses}} attribute.}
	\label{fig:supp-faces-eyeglasses}
\end{figure}

\begin{figure}[htb]
	\centering
	\includegraphics[width=0.9\linewidth]{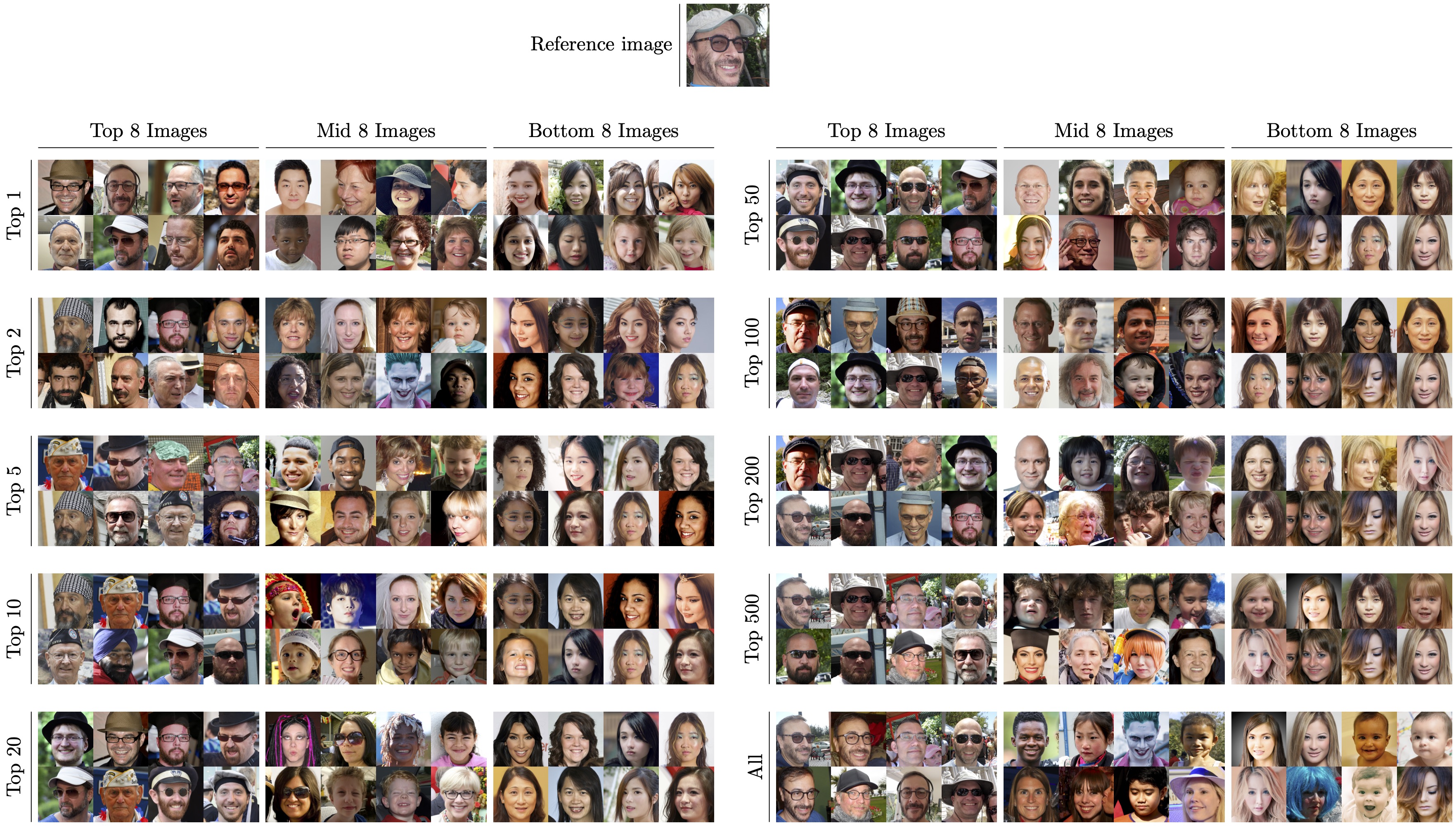}
	\caption{Ranked matches to the reference image using \texttt{dna-emd} with neurons of the Mugs (ViT-B/16) feature extractor sensitive to the \textbf{\texttt{wearing hat}} attribute.}
	\label{fig:supp-faces-hat}
\end{figure}

In \cref{fig:supp-faces-eyeglasses,fig:supp-faces-hat}, we present ranked matches using different numbers of selected neurons to focus on specific attributes.
For the \texttt{wearing hat} attribute, we see that using too few or too many neurons can lead to not focusing on the desired attribute anymore.
For the \texttt{eyeglasses} attribute, we are able to focus on the correct matches with all numbers of neurons. Even when using all neurons and no selection strategy, images with the attribute seem sufficiently favoured to be better ranked.

\subsection{StyleGANv2 ranked images}

\begin{figure}[htb]
	\centering
	\includegraphics[width=0.9\linewidth]{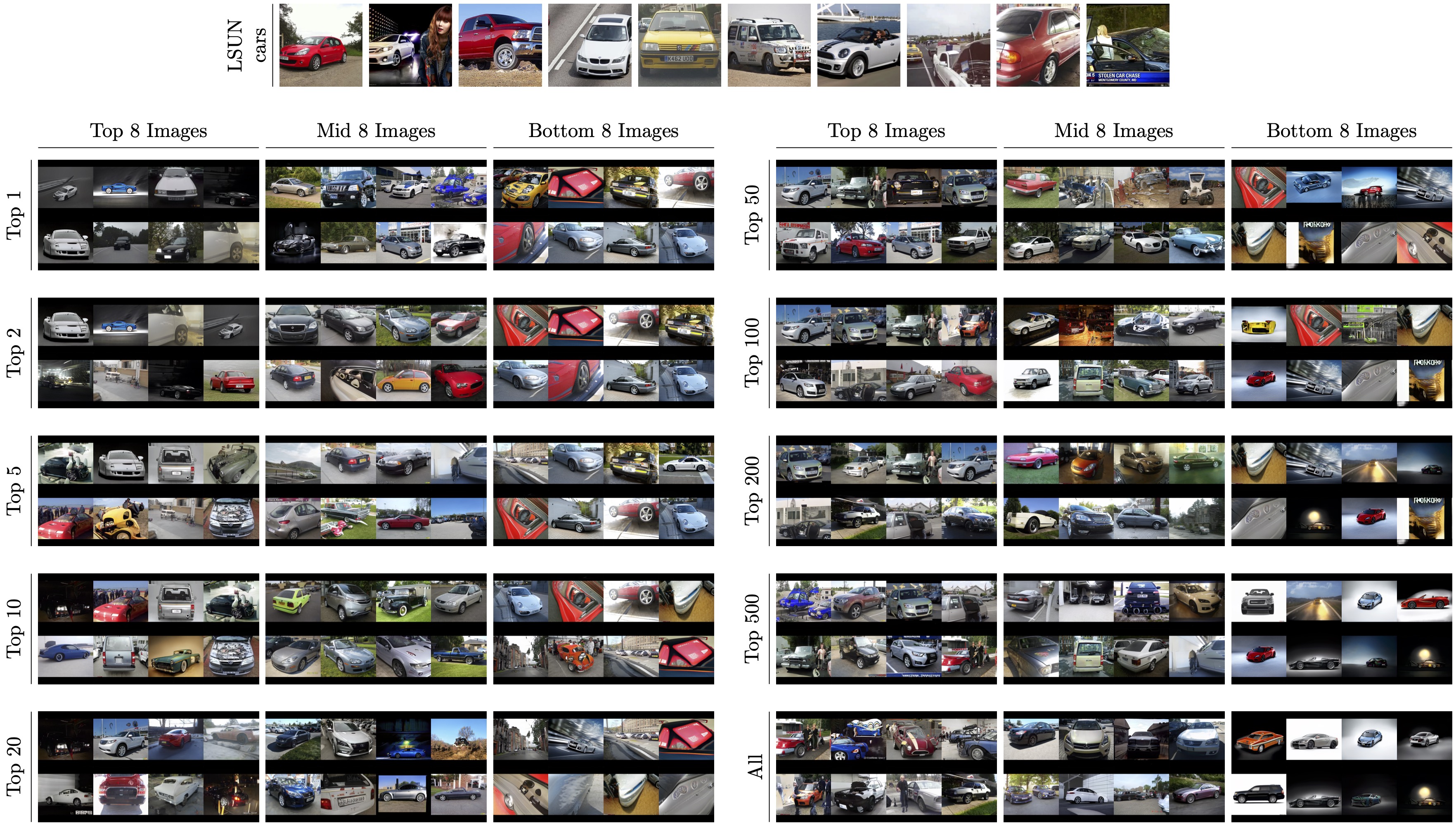}
	\caption{Generated StyleGANv2~\cite{karrasAnalyzingImprovingImage2020a} car images ranked by \texttt{dna-emd} when compared to the LSUN car images. We show the rankings for different numbers of selected neurons. The neuron selection strategy selects the most sensitive neurons when comparing real and synthetic images from other datasets.}
	\label{fig:supp-cars}
\end{figure}

\begin{figure}[htb]
	\centering
	\includegraphics[width=0.9\linewidth]{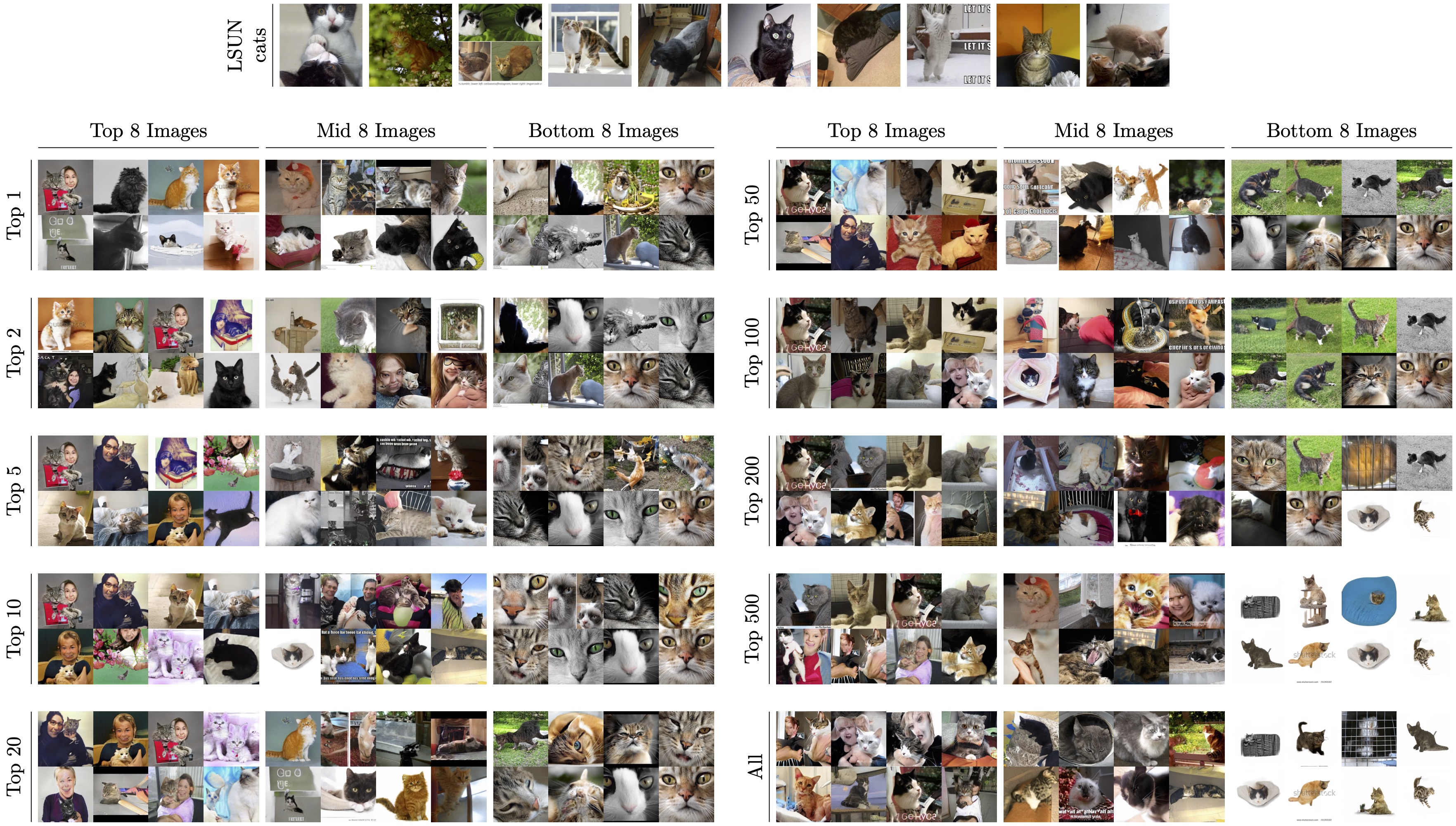}
	\caption{Generated StyleGANv2~\cite{karrasAnalyzingImprovingImage2020a} cat images ranked by \texttt{dna-emd} when compared to the LSUN cat images. We show the rankings for different numbers of selected neurons. The neuron selection strategy selects the most sensitive neurons when comparing real and synthetic images from other datasets.}
	\label{fig:supp-cats}
\end{figure}

\begin{figure}[htb]
	\centering
	\includegraphics[width=0.9\linewidth]{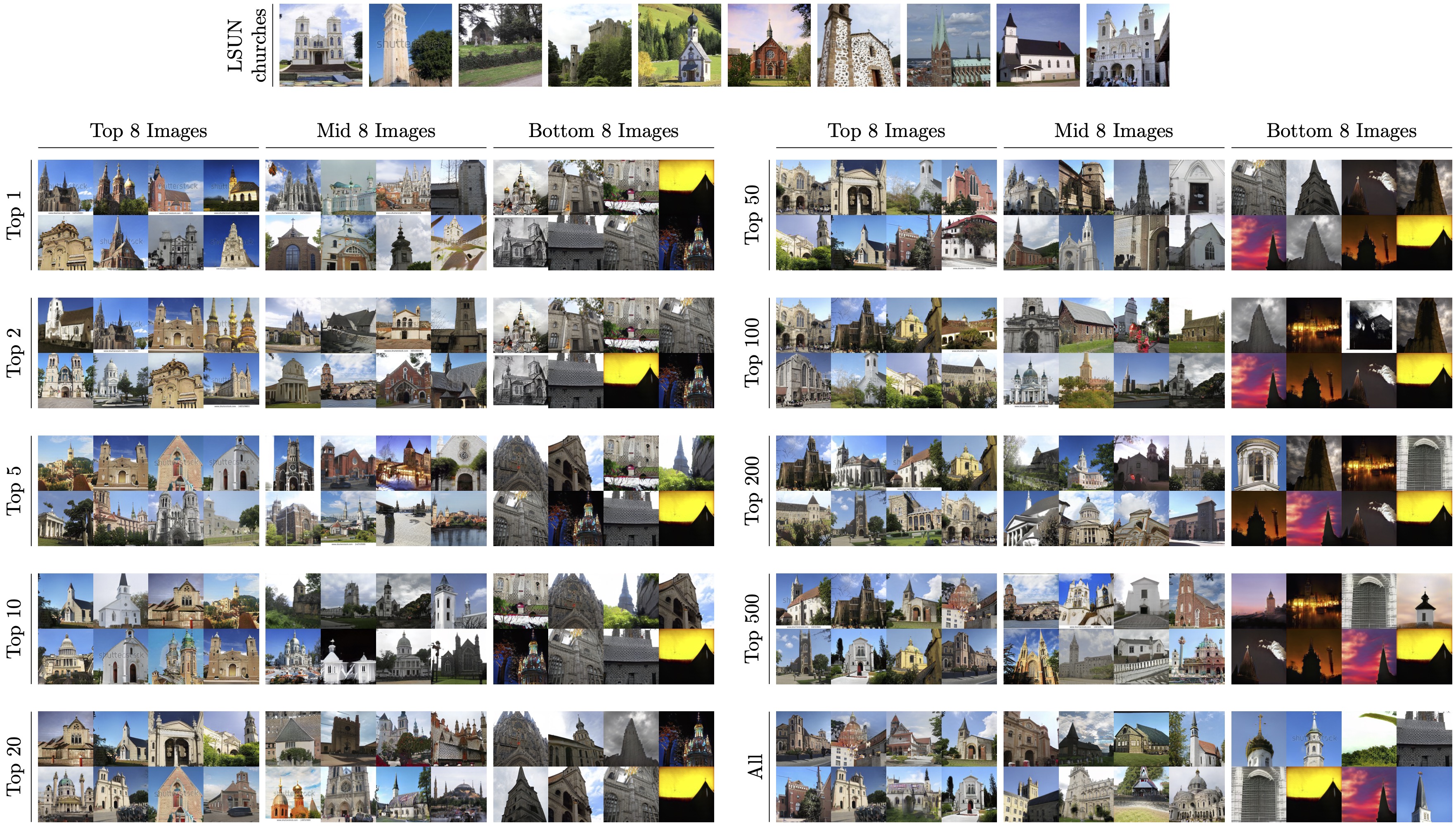}
	\caption{Generated StyleGANv2~\cite{karrasAnalyzingImprovingImage2020a} church images ranked by \texttt{dna-emd} when compared to the LSUN church images. We show the rankings for different numbers of selected neurons. The neuron selection strategy selects the most sensitive neurons when comparing real and synthetic images from other datasets.}
	\label{fig:supp-churches}
\end{figure}

\begin{figure}[htb]
	\centering
	\includegraphics[width=0.9\linewidth]{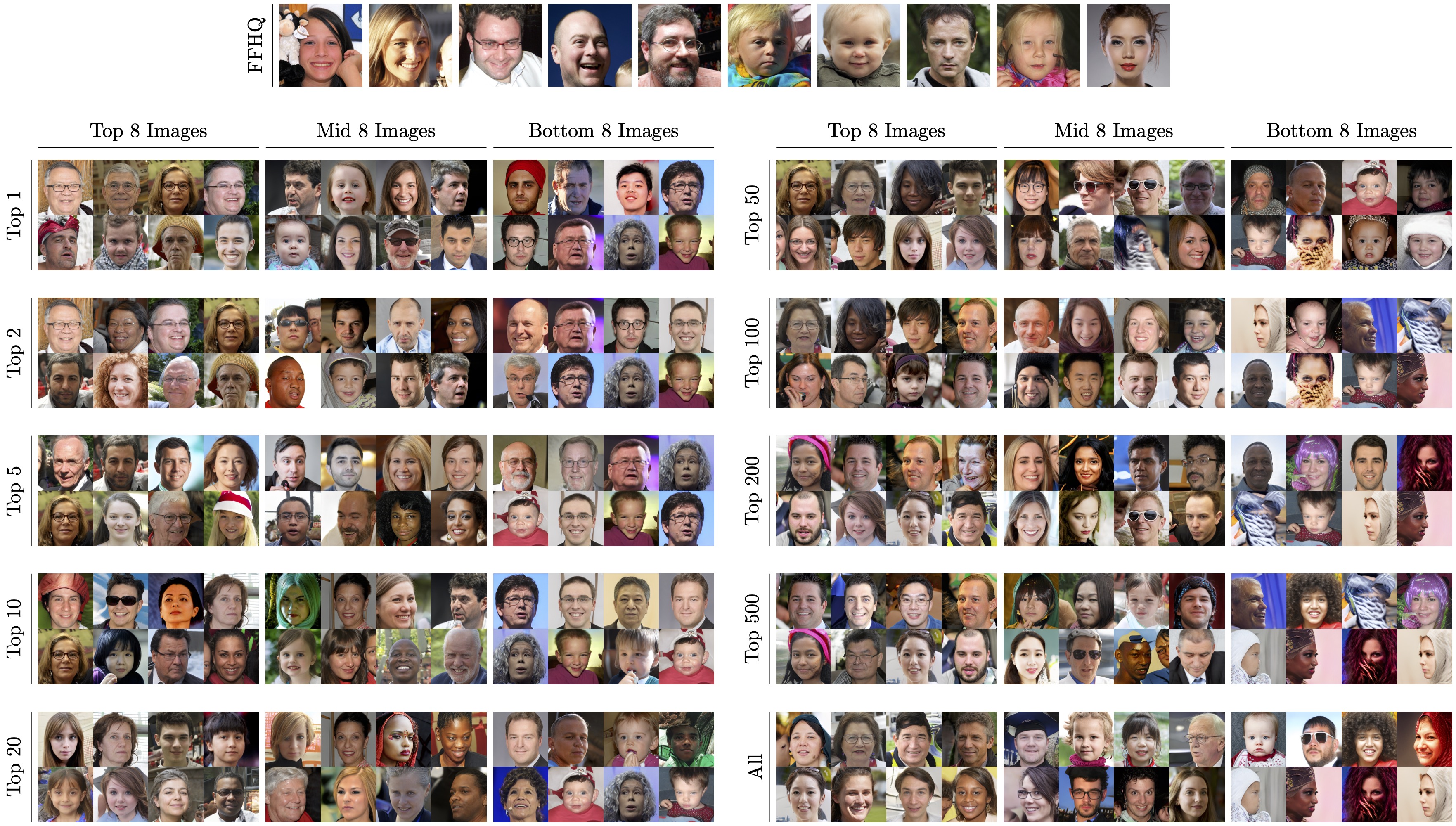}
	\caption{Generated StyleGANv2~\cite{karrasAnalyzingImprovingImage2020a} face images ranked by \texttt{dna-emd} when compared to the FFHQ images. We show the rankings for different numbers of selected neurons. The neuron selection strategy selects the most sensitive neurons when comparing real and synthetic images from other datasets.}
	\label{fig:supp-ffhq}
\end{figure}

\begin{figure}[htb]
	\centering
	\includegraphics[width=0.9\linewidth]{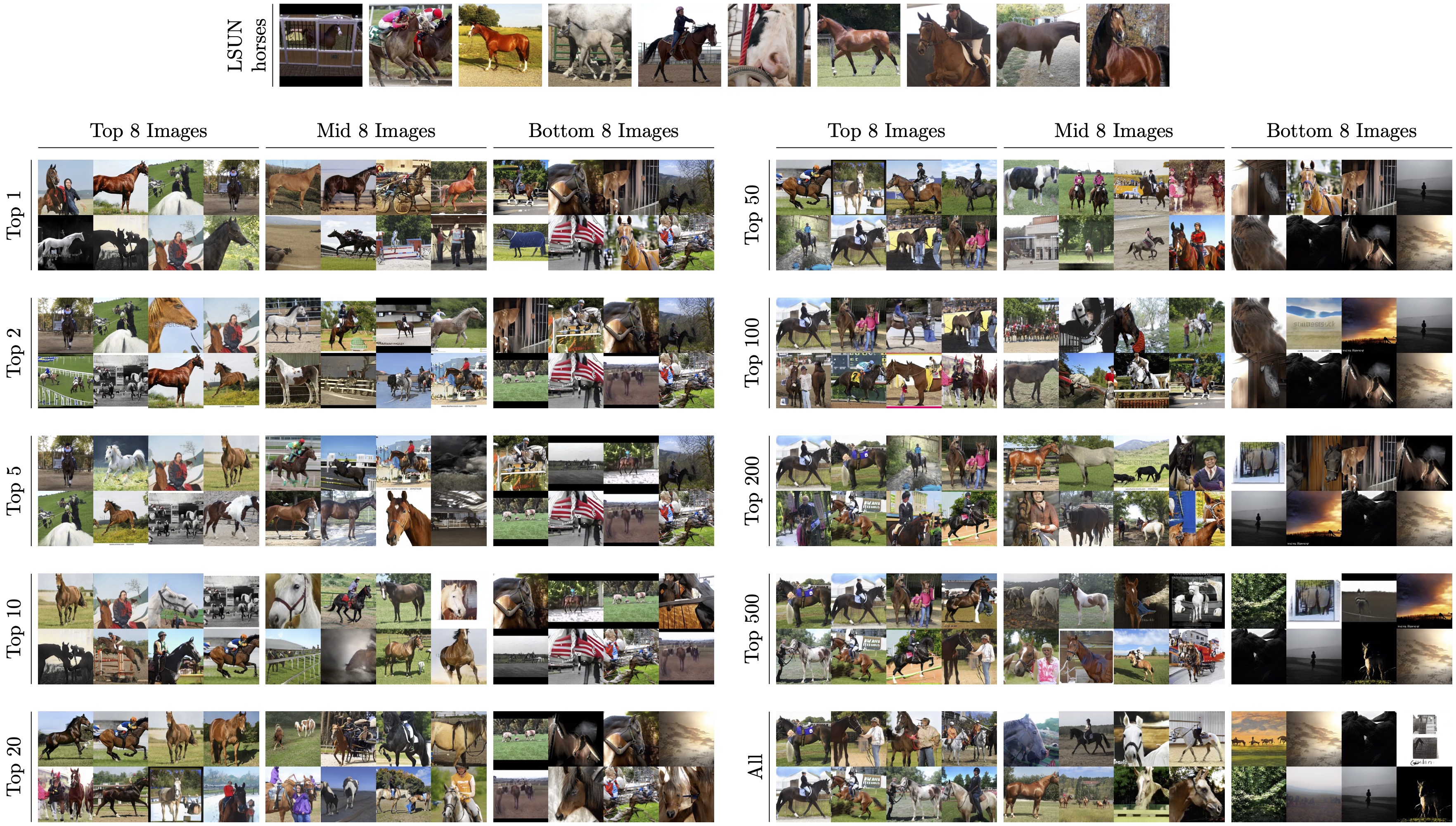}
	\caption{Generated StyleGANv2~\cite{karrasAnalyzingImprovingImage2020a} horse images ranked by \texttt{dna-emd} when compared to the LSUN horse images. We show the rankings for different numbers of selected neurons. The neuron selection strategy selects the most sensitive neurons when comparing real and synthetic images from other datasets.}
	\label{fig:supp-horses}
\end{figure}

We present more results of synthetic StyleGANv2 image rankings for different numbers of selected neurons for realism in \cref{fig:supp-cars,fig:supp-cats,fig:supp-churches,fig:supp-ffhq,fig:supp-horses}.

\subsection{Cross-dataset generalisation}
Finally, in \cref{tab:detailed-mseg-ours,tab:detailed-mseg-univ,tab:detailed-mseg-val,tab:detailed-mseg-nfd,tab:detailed-mseg-fd} we present detailed cross-dataset generalisation results that are used to produce \cref{tab:mseg-summary}. Details are provided for the HRNet-W48 and Mugs models using \texttt{dna-emd}, and for the Mugs model with \texttt{dna-fd} and \texttt{fd}.

\begin{table*}[ht]
	\centering
	\begin{subtable}[h]{\textwidth}
		\centering
		\resizebox{\columnwidth}{!}{
			\begin{tabular}{l  l  l  l  l  l  l  l}\toprule
				\multicolumn{1}{c}{}                                                        & \multicolumn{7}{c}{Validation datasets}                                                                                                                       \\
				\cmidrule(lr){2-8}
				                                                                            & ADE20K                                  & BDD100K           & Cityscapes        & COCO             & IDD               & Mapillary         & SUN-RGBD         \\
				\cmidrule(lr){2-2}\cmidrule(lr){3-3}\cmidrule(lr){4-4}\cmidrule(lr){5-5}\cmidrule(lr){6-6}\cmidrule(lr){7-7}\cmidrule(lr){8-8}
				\multirow{7}{*}{\rotatebox{90}{\parbox{3cm}{\centering Training datasets}}} & 45.3 - ADE20K                           & 63.2 - BDD100K    & 77.6 - Cityscapes & 52.6 - COCO      & 64.8 - IDD        & 56.2 - Mapillary  & 43.9 - SUN-RGBD  \\
				                                                                            & 19.6 - COCO                             & 60.2 - Mapillary  & 69.7 - Mapillary  & 14.5 - ADE20K    & 48.2 - Mapillary  & 26.7 - COCO       & 35.3 - ADE20K    \\
				                                                                            & 7.1 - SUN-RGBD                          & 45.0 - Cityscapes & 60.9 - BDD100K    & 6.7 - Mapillary  & 33.9 - BDD100K    & 24.3 - ADE20K     & 29.4 - COCO      \\
				                                                                            & 6.2 - Mapillary                         & 44.1 - COCO       & 50.2 - IDD        & 3.7 - BDD100K    & 31.3 - Cityscapes & 24.3 - IDD        & 0.6 - IDD        \\
				                                                                            & 4.1 - BDD100K                           & 43.7 - IDD        & 46.2 - COCO       & 3.3 - SUN-RGBD   & 31.0 - COCO       & 24.0 - BDD100K    & 0.2 - BDD100K    \\
				                                                                            & 3.1 - Cityscapes                        & 41.5 - ADE20K     & 44.3 - ADE20K     & 3.1 - Cityscapes & 27.0 - ADE20K     & 22.4 - Cityscapes & 0.2 - Cityscapes \\
				                                                                            & 3.1 - IDD                               & 2.2 - SUN-RGBD    & 2.6 - SUN-RGBD    & 3.1 - IDD        & 1.0 - SUN-RGBD    & 1.1 - SUN-RGBD    & 0.2 - Mapillary  \\\bottomrule
			\end{tabular}}
		\caption{Observed cross-dataset generalisation on semantic segmentation from Lambert \etal~\cite{lambertMSegCompositeDataset2020} (mIoU). Each column corresponds to the evaluation of one dataset (validation set). Rows are ordered by a cross-generalisation performance from training on each dataset (training set).}
		\label{tab:mseg_ref}
	\end{subtable}
	\newline
	\vspace*{0.4cm}
	\newline
	\begin{subtable}[h]{\textwidth}
		\centering
		\resizebox{\columnwidth}{!}{
			\begin{tabular}{l l l l l l l l }\toprule
				\multicolumn{1}{c}{}                                                          & \multicolumn{7}{c}{Validation datasets}                                                                                                                                                     \\
				\cmidrule(lr){2-8}
				                                                                              & ADE20K                                  & BDD100K                 & Cityscapes          & COCO                    & IDD                 & Mapillary               & SUN-RGBD                \\
				\cmidrule(lr){2-2}\cmidrule(lr){3-3}\cmidrule(lr){4-4}\cmidrule(lr){5-5}\cmidrule(lr){6-6}\cmidrule(lr){7-7}\cmidrule(lr){8-8}
				\multirow{7}{*}{\rotatebox{90}{\parbox{2.9cm}{\centering Training datasets}}} & 0.84 - ADE20K                           & 3.61 - BDD100K          & 3.18 - Cityscapes   & 0.64 - COCO             & 4.48 - IDD          & 0.85 - Mapillary        & 2.22 - SUN-RGBD         \\
				                                                                              & 9.17 - COCO                             & 8.91 - Mapillary        & 15.9 - Mapillary    & 9.2 - ADE20K            & 12.88 - Mapillary   & 9.72 - BDD100K     & 12.71 - ADE20K          \\
				                                                                              & 12.66 - SUN-RGBD                        & 13.03 - IDD        & 17.13 - BDD100K     & 15.4 - SUN-RGBD    & 13.54 - BDD100K     & 11.52 - IDD        & 15.57 - COCO            \\
				                                                                              & 21.18 - Mapillary                       & 17.18 - Cityscapes & 17.15 - IDD         & 21.79 - Mapillary  & 18.95 - Cityscapes  & 16.19 - Cityscapes & 26.92 - IDD             \\
				                                                                              & 21.57 - IDD                        & 22.56 - ADE20K     & 25.61 - ADE20K & 22.36 - IDD        & 22.53 - ADE20K & 21.26 - ADE20K     & 27.01 - Mapillary  \\
				                                                                              & 22.58 - BDD100K                    & 23.88 - COCO       & 25.73 - COCO   & 23.86 - BDD100K    & 23.33 - COCO  & 21.89 - COCO       & 27.96 - BDD100K    \\
				                                                                              & 26.17 - Cityscapes                 & 27.96 - SUN-RGBD        & 30.47 - SUN-RGBD    & 26.41 - Cityscapes & 27.51 - SUN-RGBD    & 26.95 - SUN-RGBD        & 31.08 - Cityscapes \\\bottomrule
			\end{tabular}}
		\caption{Ordered datasets ranked by \texttt{dna-emd} with corresponding EMD values. The EMD here is computed using the last layer of the Mugs (ViT-B/16) feature extractor.}
		\label{tab:mseg_ours_dists}
	\end{subtable}
	\newline
	\vspace*{0.4 cm}
	\newline
	\begin{subtable}[h]{\textwidth}
		\centering
		\resizebox{\columnwidth}{!}{
			\begin{tabular}{l l l l l l l l }\toprule
				\multicolumn{1}{c}{}                                                          & \multicolumn{7}{c}{Validation datasets}                                                                                                                       \\
				\cmidrule(lr){2-8}
				                                                                              & ADE20K                                  & BDD100K           & Cityscapes        & COCO             & IDD               & Mapillary         & SUN-RGBD         \\
				\cmidrule(lr){2-2}\cmidrule(lr){3-3}\cmidrule(lr){4-4}\cmidrule(lr){5-5}\cmidrule(lr){6-6}\cmidrule(lr){7-7}\cmidrule(lr){8-8}
				\multirow{7}{*}{\rotatebox{90}{\parbox{2.9cm}{\centering Training datasets}}} & 45.3 - ADE20K                           & 63.2 - BDD100K    & 77.6 - Cityscapes & 52.6 - COCO      & 64.8 - IDD        & 56.2 - Mapillary  & 43.9 - SUN-RGBD  \\
				                                                                              & 19.6 - COCO                             & 60.2 - Mapillary  & 69.7 - Mapillary  & 14.5 - ADE20K    & 48.2 - Mapillary  & 24.0 - BDD100K    & 35.3 - ADE20K    \\
				                                                                              & 7.1 - SUN-RGBD                          & 43.7 - IDD        & 60.9 - BDD100K    & 3.3 - SUN-RGBD   & 33.9 - BDD100K    & 24.3 - IDD        & 29.4 - COCO      \\
				                                                                              & 6.2 - Mapillary                         & 45.0 - Cityscapes & 50.2 - IDD        & 6.7 - Mapillary  & 31.3 - Cityscapes & 22.4 - Cityscapes & 0.6 - IDD        \\
				                                                                              & 3.1 - IDD                               & 41.5 - ADE20K     & 44.3 - ADE20K     & 3.1 - IDD        & 27.0 - ADE20K     & 24.3 - ADE20K     & 0.2 - Mapillary  \\
				                                                                              & 4.1 - BDD100K                           & 44.1 - COCO       & 46.2 - COCO       & 3.7 - BDD100K    & 31.0 - COCO       & 26.7 - COCO       & 0.2 - BDD100K    \\
				                                                                              & 3.1 - Cityscapes                        & 2.2 - SUN-RGBD    & 2.6 - SUN-RGBD    & 3.1 - Cityscapes & 1.0 - SUN-RGBD    & 1.1 - SUN-RGBD    & 0.2 - Cityscapes \\\bottomrule
			\end{tabular}}
		\caption{Ordered datasets ranked by \texttt{dna-emd} with corresponding mIoU values. The ranking is taken from \cref{tab:mseg_ours_dists}, but we show the mIoUs for the corresponding datasets from \cref{tab:mseg_ref} instead.}
		\label{tab:mseg_ours_miou}
	\end{subtable}
	\newline
	\vspace*{0.4 cm}
	\newline
	\begin{subtable}[h]{\textwidth}
 \scriptsize
		\centering
		\resizebox{0.65\columnwidth}{!}{
			\begin{tabular}{c c c c c c c }\toprule
				\multicolumn{7}{c}{Validation datasets}                           \\
				\cmidrule(lr){1-7}
				ADE20K & BDD100K & Cityscapes & COCO & IDD & Mapillary & SUN-RGBD \\
				\cmidrule(lr){1-1}\cmidrule(lr){2-2}\cmidrule(lr){3-3}\cmidrule(lr){4-4}\cmidrule(lr){5-5}\cmidrule(lr){6-6}\cmidrule(lr){7-7}
				0.0    & 0.0     & 0.0        & 0.0  & 0.0 & 0.0       & 0.0      \\
				0.0    & 0.0     & 0.0        & 0.0  & 0.0 & 2.7       & 0.0      \\
				0.0    & 1.3     & 0.0        & 3.4  & 0.0 & 0.0       & 0.0      \\
				0.0    & 0.9     & 0.0        & 3.0  & 0.0 & 1.9       & 0.0      \\
				1.0    & 2.2     & 1.9        & 0.2  & 4.0 & 0.3       & 0.0      \\
				1.0    & 2.6     & 1.9        & 0.6  & 4.0 & 4.3       & 0.0      \\
				0.0    & 0.0     & 0.0        & 0.0  & 0.0 & 0.0       & 0.0      \\\cmidrule(lr){1-7}
				\multicolumn{7}{c}{Average absolute mIoU difference: 0.76}        \\\bottomrule
			\end{tabular}
		}
		\caption{Differences between the mIoUs of the reference ranking (\cref{tab:mseg_ref}) and the mIoUs for the predicted \texttt{dna-emd} ranking (\cref{tab:mseg_ours_miou}).}
		\label{tab:mseg_diffs}
	\end{subtable}
	\newline
	\caption{Detailed results comparing the observed cross-dataset generalisation of the HRNet-W48 semantic segmentation models to predictions relying only on datasets using \textbf{\texttt{dna-emd}} with the \textbf{Mugs (ViT-B/16) feature extractor}. The final value reported in \cref{tab:mseg-summary} corresponds to the average of the values in \cref{tab:mseg_diffs}.}
	\label{tab:detailed-mseg-ours}
\end{table*}

\begin{table*}[ht]
	\centering
	\begin{subtable}[h]{\textwidth}
		\centering
		\resizebox{\columnwidth}{!}{
			\begin{tabular}{l  l  l  l  l  l  l  l}\toprule
				\multicolumn{1}{c}{}                                                        & \multicolumn{7}{c}{Validation datasets}                                                                                                                       \\
				\cmidrule(lr){2-8}
				                                                                            & ADE20K                                  & BDD100K           & Cityscapes        & COCO             & IDD               & Mapillary         & SUN-RGBD         \\
				\cmidrule(lr){2-2}\cmidrule(lr){3-3}\cmidrule(lr){4-4}\cmidrule(lr){5-5}\cmidrule(lr){6-6}\cmidrule(lr){7-7}\cmidrule(lr){8-8}
				\multirow{7}{*}{\rotatebox{90}{\parbox{3cm}{\centering Training datasets}}} & 45.3 - ADE20K                           & 63.2 - BDD100K    & 77.6 - Cityscapes & 52.6 - COCO      & 64.8 - IDD        & 56.2 - Mapillary  & 43.9 - SUN-RGBD  \\
				                                                                            & 19.6 - COCO                             & 60.2 - Mapillary  & 69.7 - Mapillary  & 14.5 - ADE20K    & 48.2 - Mapillary  & 26.7 - COCO       & 35.3 - ADE20K    \\
				                                                                            & 7.1 - SUN-RGBD                          & 45.0 - Cityscapes & 60.9 - BDD100K    & 6.7 - Mapillary  & 33.9 - BDD100K    & 24.3 - ADE20K     & 29.4 - COCO      \\
				                                                                            & 6.2 - Mapillary                         & 44.1 - COCO       & 50.2 - IDD        & 3.7 - BDD100K    & 31.3 - Cityscapes & 24.3 - IDD        & 0.6 - IDD        \\
				                                                                            & 4.1 - BDD100K                           & 43.7 - IDD        & 46.2 - COCO       & 3.3 - SUN-RGBD   & 31.0 - COCO       & 24.0 - BDD100K    & 0.2 - BDD100K    \\
				                                                                            & 3.1 - Cityscapes                        & 41.5 - ADE20K     & 44.3 - ADE20K     & 3.1 - Cityscapes & 27.0 - ADE20K     & 22.4 - Cityscapes & 0.2 - Cityscapes \\
				                                                                            & 3.1 - IDD                               & 2.2 - SUN-RGBD    & 2.6 - SUN-RGBD    & 3.1 - IDD        & 1.0 - SUN-RGBD    & 1.1 - SUN-RGBD    & 0.2 - Mapillary  \\\bottomrule
			\end{tabular}}
		\caption{Observed cross-dataset generalisation on semantic segmentation from Lambert \etal~\cite{lambertMSegCompositeDataset2020} (mIoU). Each column corresponds to the evaluation of one dataset (validation set). Rows are ordered by a cross-generalisation performance from training on each dataset (training set).}
		\label{tab:mseg_ref_univ}
	\end{subtable}
	\newline
	\vspace*{0.4cm}
	\newline
	\begin{subtable}[h]{\textwidth}
		\centering
		\resizebox{\columnwidth}{!}{
			\begin{tabular}{l l l l l l l l }\toprule
				\multicolumn{1}{c}{}                                                          & \multicolumn{7}{c}{Validation datasets}                                                                                                                         \\
				\cmidrule(lr){2-8}
				                                                                              & ADE20K                                  & BDD100K           & Cityscapes        & COCO              & IDD               & Mapillary         & SUN-RGBD          \\
				\cmidrule(lr){2-2}\cmidrule(lr){3-3}\cmidrule(lr){4-4}\cmidrule(lr){5-5}\cmidrule(lr){6-6}\cmidrule(lr){7-7}\cmidrule(lr){8-8}
				\multirow{7}{*}{\rotatebox{90}{\parbox{2.9cm}{\centering Training datasets}}} & 0.02 - ADE20K                           & 0.08 - BDD100K    & 0.04 - Cityscapes & 0.01 - COCO       & 0.07 - IDD        & 0.02 - Mapillary  & 0.03 - SUN-RGBD   \\
				                                                                              & 0.3 - SUN-RGBD                          & 0.31 - IDD        & 0.35 - Mapillary  & 0.27 - Mapillary  & 0.32 - Mapillary  & 0.26 - COCO       & 0.3 - ADE20K      \\
				                                                                              & 0.32 - BDD100K                          & 0.34 - SUN-RGBD   & 0.37 - COCO       & 0.31 - IDD        & 0.32 - COCO       & 0.33 - IDD        & 0.31 - BDD100K    \\
				                                                                              & 0.51 - IDD                              & 0.34 - ADE20K     & 0.46 - IDD        & 0.37 - Cityscapes & 0.35 - BDD100K    & 0.34 - Cityscapes & 0.47 - IDD        \\
				                                                                              & 0.6 - COCO                              & 0.51 - Mapillary  & 0.7 - BDD100K     & 0.54 - SUN-RGBD   & 0.45 - Cityscapes & 0.56 - BDD100K    & 0.54 - COCO       \\
				                                                                              & 0.68 - Mapillary                        & 0.52 - COCO       & 0.75 - SUN-RGBD   & 0.56 - BDD100K    & 0.47 - SUN-RGBD   & 0.63 - SUN-RGBD   & 0.63 - Mapillary  \\
				                                                                              & 0.81 - Cityscapes                       & 0.66 - Cityscapes & 0.82 - ADE20K     & 0.61 - ADE20K     & 0.52 - ADE20K     & 0.7 - ADE20K      & 0.75 - Cityscapes \\\bottomrule
			\end{tabular}}
		\caption{Ordered datasets ranked by \texttt{dna-emd} with corresponding EMD values. The EMD here is computed using the last layer of the HRNet-W48 feature extractor trained on all domains.}
		\label{tab:mseg_univ_dists}
	\end{subtable}
	\newline
	\vspace*{0.4 cm}
	\newline
	\begin{subtable}[h]{\textwidth}
		\centering
		\resizebox{\columnwidth}{!}{
			\begin{tabular}{l l l l l l l l }\toprule
				\multicolumn{1}{c}{}                                                          & \multicolumn{7}{c}{Validation datasets}                                                                                                                      \\
				\cmidrule(lr){2-8}
				                                                                              & ADE20K                                  & BDD100K           & Cityscapes        & COCO             & IDD               & Mapillary         & SUN-RGBD        \\
				\cmidrule(lr){2-2}\cmidrule(lr){3-3}\cmidrule(lr){4-4}\cmidrule(lr){5-5}\cmidrule(lr){6-6}\cmidrule(lr){7-7}\cmidrule(lr){8-8}
				\multirow{7}{*}{\rotatebox{90}{\parbox{2.9cm}{\centering Training datasets}}} & 45.3 - ADE20K                           & 63.2 - BDD100K    & 77.6 - Cityscapes & 52.6 - COCO      & 64.8 - IDD        & 56.2 - Mapillary  & 43.9 - SUN-RGBD \\
				                                                                              & 7.1 - SUN-RGBD                          & 43.7 - IDD        & 69.7 - Mapillary  & 6.7 - Mapillary  & 48.2 - Mapillary  & 26.7 - COCO       & 35.3 - ADE20K   \\
				                                                                              & 4.1 - BDD100K                           & 2.2 - SUN-RGBD    & 46.2 - COCO       & 3.1 - IDD        & 31.0 - COCO       & 24.3 - IDD        & 0.2 - BDD100K   \\
				                                                                              & 3.1 - IDD                               & 41.5 - ADE20K     & 50.2 - IDD        & 3.1 - Cityscapes & 33.9 - BDD100K    & 22.4 - Cityscapes & 0.6 - IDD       \\
				                                                                              & 19.6 - COCO                             & 60.2 - Mapillary  & 60.9 - BDD100K    & 3.3 - SUN-RGBD   & 31.3 - Cityscapes & 24.0 - BDD100K    & 29.4 - COCO     \\
				                                                                              & 6.2 - Mapillary                         & 44.1 - COCO       & 2.6 - SUN-RGBD    & 3.7 - BDD100K    & 1.0 - SUN-RGBD    & 1.1 - SUN-RGBD    & 0.2 - Mapillary \\
				                                                                              & 3.1 - Cityscapes                        & 45.0 - Cityscapes & 44.3 - ADE20K     & 14.5 - ADE20K    & 27.0 - ADE20K     & 24.3 - ADE20K     & 0.2 - Cityscape \\\bottomrule
			\end{tabular}}
		\caption{Ordered datasets ranked by \texttt{dna-emd} with corresponding mIoU values. The ranking is taken from \cref{tab:mseg_univ_dists}, but we show the mIoUs for the corresponding datasets from \cref{tab:mseg_ref_univ} instead.}
		\label{tab:mseg_univ_miou}
	\end{subtable}
	\newline
	\vspace*{0.4 cm}
	\newline
	\begin{subtable}[h]{\textwidth}
	 \scriptsize
		\centering
		\resizebox{0.65\columnwidth}{!}{
			\begin{tabular}{c c c c c c c }\toprule
				\multicolumn{7}{c}{Validation datasets}                            \\
				\cmidrule(lr){1-7}
				ADE20K & BDD100K & Cityscapes & COCO & IDD  & Mapillary & SUN-RGBD \\
				\cmidrule(lr){1-1}\cmidrule(lr){2-2}\cmidrule(lr){3-3}\cmidrule(lr){4-4}\cmidrule(lr){5-5}\cmidrule(lr){6-6}\cmidrule(lr){7-7}
				0.0    & 0.0     & 0.0        & 0.0  & 0.0  & 0.0       & 0.0      \\
				12.5   & 16.5    & 0.0        & 7.8  & 0.0  & 0.0       & 0.0      \\
				3.0    & 42.8    & 14.7       & 3.6  & 2.9  & 0.0       & 29.2     \\
				3.1    & 2.6     & 0.0        & 0.6  & 2.6  & 1.9       & 0.0      \\
				15.5   & 16.5    & 14.7       & 0.0  & 0.3  & 0.0       & 29.2     \\
				3.1    & 2.6     & 41.7       & 0.6  & 26.0 & 21.3      & 0.0      \\
				0.0    & 42.8    & 41.7       & 11.4 & 26.0 & 23.2      & 0.0      \\\cmidrule(lr){1-7}
				\multicolumn{7}{c}{Average absolute mIoU difference: 9.40}         \\\bottomrule
			\end{tabular}
		}
		\caption{Differences between the mIoUs of the reference ranking (\cref{tab:mseg_ref_univ}) and the mIoUs for the predicted \texttt{dna-emd} ranking (\cref{tab:mseg_univ_miou}).}
		\label{tab:mseg_univ_diffs}
	\end{subtable}
	\newline
	\caption{Detailed results comparing the observed cross-dataset generalisation of the HRNet-W48 semantic segmentation models to predictions relying only on datasets using \textbf{\texttt{dna-emd}} with the \textbf{HRNet-W48 feature extractor trained on all domains}. The final value reported in \cref{tab:mseg-summary} corresponds to the average of the values in \cref{tab:mseg_univ_diffs}.}
	\label{tab:detailed-mseg-univ}
\end{table*}

\begin{table*}[ht]
	\centering
	\begin{subtable}[h]{\textwidth}
		\centering
		\resizebox{\columnwidth}{!}{
			\begin{tabular}{l  l  l  l  l  l  l  l}\toprule
				\multicolumn{1}{c}{}                                                        & \multicolumn{7}{c}{Validation datasets}                                                                                                                       \\
				\cmidrule(lr){2-8}
				                                                                            & ADE20K                                  & BDD100K           & Cityscapes        & COCO             & IDD               & Mapillary         & SUN-RGBD         \\
				\cmidrule(lr){2-2}\cmidrule(lr){3-3}\cmidrule(lr){4-4}\cmidrule(lr){5-5}\cmidrule(lr){6-6}\cmidrule(lr){7-7}\cmidrule(lr){8-8}
				\multirow{7}{*}{\rotatebox{90}{\parbox{3cm}{\centering Training datasets}}} & 45.3 - ADE20K                           & 63.2 - BDD100K    & 77.6 - Cityscapes & 52.6 - COCO      & 64.8 - IDD        & 56.2 - Mapillary  & 43.9 - SUN-RGBD  \\
				                                                                            & 19.6 - COCO                             & 60.2 - Mapillary  & 69.7 - Mapillary  & 14.5 - ADE20K    & 48.2 - Mapillary  & 26.7 - COCO       & 35.3 - ADE20K    \\
				                                                                            & 7.1 - SUN-RGBD                          & 45.0 - Cityscapes & 60.9 - BDD100K    & 6.7 - Mapillary  & 33.9 - BDD100K    & 24.3 - ADE20K     & 29.4 - COCO      \\
				                                                                            & 6.2 - Mapillary                         & 44.1 - COCO       & 50.2 - IDD        & 3.7 - BDD100K    & 31.3 - Cityscapes & 24.3 - IDD        & 0.6 - IDD        \\
				                                                                            & 4.1 - BDD100K                           & 43.7 - IDD        & 46.2 - COCO       & 3.3 - SUN-RGBD   & 31.0 - COCO       & 24.0 - BDD100K    & 0.2 - BDD100K    \\
				                                                                            & 3.1 - Cityscapes                        & 41.5 - ADE20K     & 44.3 - ADE20K     & 3.1 - Cityscapes & 27.0 - ADE20K     & 22.4 - Cityscapes & 0.2 - Cityscapes \\
				                                                                            & 3.1 - IDD                               & 2.2 - SUN-RGBD    & 2.6 - SUN-RGBD    & 3.1 - IDD        & 1.0 - SUN-RGBD    & 1.1 - SUN-RGBD    & 0.2 - Mapillary  \\\bottomrule
			\end{tabular}}
		\caption{Observed cross-dataset generalisation on semantic segmentation from Lambert \etal~\cite{lambertMSegCompositeDataset2020} (mIoU). Each column corresponds to the evaluation of one dataset (validation set). Rows are ordered by a cross-generalisation performance from training on each dataset (training set).}
		\label{tab:mseg_ref_val}
	\end{subtable}
	\newline
	\vspace*{0.4cm}
	\newline
	\begin{subtable}[h]{\textwidth}
		\centering
		\resizebox{\columnwidth}{!}{
			\begin{tabular}{l l l l l l l l }\toprule
				\multicolumn{1}{c}{}                                                          & \multicolumn{7}{c}{Validation datasets}                                                                                                                        \\
				\cmidrule(lr){2-8}
				                                                                              & ADE20K                                  & BDD100K           & Cityscapes        & COCO              & IDD              & Mapillary         & SUN-RGBD          \\
				\cmidrule(lr){2-2}\cmidrule(lr){3-3}\cmidrule(lr){4-4}\cmidrule(lr){5-5}\cmidrule(lr){6-6}\cmidrule(lr){7-7}\cmidrule(lr){8-8}
				\multirow{7}{*}{\rotatebox{90}{\parbox{2.9cm}{\centering Training datasets}}} & 0.01 - ADE20K                           & 0.04 - BDD100K    & 0.02 - Cityscapes & 0.0 - COCO        & 0.02 - IDD       & 0.0 - Mapillary   & 0.03 - SUN-RGBD   \\
				                                                                              & 0.1 - COCO                              & 0.11 - IDD        & 0.15 - IDD        & 0.1 - ADE20K      & 0.08 - BDD100K   & 0.04 - BDD100K    & 0.17 - COCO       \\
				                                                                              & 0.14 - SUN-RGBD                         & 0.11 - Mapillary  & 0.19 - BDD100K    & 0.12 - SUN-RGBD   & 0.1 - Cityscapes & 0.05 - IDD        & 0.17 - IDD        \\
				                                                                              & 0.21 - BDD100K                          & 0.15 - Cityscapes & 0.24 - Mapillary  & 0.13 - BDD100K    & 0.1 - Mapillary  & 0.06 - ADE20K     & 0.19 - ADE20K     \\
				                                                                              & 0.23 - IDD                              & 0.18 - ADE20K     & 0.31 - SUN-RGBD   & 0.16 - Mapillary  & 0.11 - ADE20K    & 0.07 - COCO       & 0.21 - Mapillary  \\
				                                                                              & 0.24 - Mapillary                        & 0.24 - COCO       & 0.34 - ADE20K     & 0.18 - IDD        & 0.11 - COCO      & 0.09 - SUN-RGBD   & 0.22 - BDD100K    \\
				                                                                              & 0.37 - Cityscapes                       & 0.28 - SUN-RGBD   & 0.37 - COCO       & 0.21 - Cityscapes & 0.12 - SUN-RGBD  & 0.12 - Cityscapes & 0.27 - Cityscapes \\\bottomrule
			\end{tabular}}
		\caption{Ordered datasets ranked by \texttt{dna-emd} with corresponding EMD values. The EMD here is computed using the last layer of the HRNet-W48 feature extractor trained on validation domains.}
		\label{tab:mseg_val_dists}
	\end{subtable}
	\newline
	\vspace*{0.4 cm}
	\newline
	\begin{subtable}[h]{\textwidth}
		\centering
		\resizebox{\columnwidth}{!}{
			\begin{tabular}{l l l l l l l l }\toprule
				\multicolumn{1}{c}{}                                                          & \multicolumn{7}{c}{Validation datasets}                                                                                                                       \\
				\cmidrule(lr){2-8}
				                                                                              & ADE20K                                  & BDD100K           & Cityscapes        & COCO             & IDD               & Mapillary         & SUN-RGBD         \\
				\cmidrule(lr){2-2}\cmidrule(lr){3-3}\cmidrule(lr){4-4}\cmidrule(lr){5-5}\cmidrule(lr){6-6}\cmidrule(lr){7-7}\cmidrule(lr){8-8}
				\multirow{7}{*}{\rotatebox{90}{\parbox{2.9cm}{\centering Training datasets}}} & 45.3 - ADE20K                           & 63.2 - BDD100K    & 77.6 - Cityscapes & 52.6 - COCO      & 64.8 - IDD        & 56.2 - Mapillary  & 43.9 - SUN-RGBD  \\
				                                                                              & 19.6 - COCO                             & 43.7 - IDD        & 50.2 - IDD        & 14.5 - ADE20K    & 33.9 - BDD100K    & 24.0 - BDD100K    & 29.4 - COCO      \\
				                                                                              & 7.1 - SUN-RGBD                          & 60.2 - Mapillary  & 60.9 - BDD100K    & 3.3 - SUN-RGBD   & 31.3 - Cityscapes & 24.3 - IDD        & 0.6 - IDD        \\
				                                                                              & 4.1 - BDD100K                           & 45.0 - Cityscapes & 69.7 - Mapillary  & 3.7 - BDD100K    & 48.2 - Mapillary  & 24.3 - ADE20K     & 35.3 - ADE20K    \\
				                                                                              & 3.1 - IDD                               & 41.5 - ADE20K     & 2.6 - SUN-RGBD    & 6.7 - Mapillary  & 27.0 - ADE20K     & 26.7 - COCO       & 0.2 - Mapillary  \\
				                                                                              & 6.2 - Mapillary                         & 44.1 - COCO       & 44.3 - ADE20K     & 3.1 - IDD        & 31.0 - COCO       & 1.1 - SUN-RGBD    & 0.2 - BDD100K    \\
				                                                                              & 3.1 - Cityscapes                        & 2.2 - SUN-RGBD    & 46.2 - COCO       & 3.1 - Cityscapes & 1.0 - SUN-RGBD    & 22.4 - Cityscapes & 0.2 - Cityscapes \\\bottomrule
			\end{tabular}}
		\caption{Ordered datasets ranked by \texttt{dna-emd} with corresponding mIoU values. The ranking is taken from \cref{tab:mseg_val_dists}, but we show the mIoUs for the corresponding datasets from \cref{tab:mseg_ref_val} instead.}
		\label{tab:mseg_val_miou}
	\end{subtable}
	\newline
	\vspace*{0.4 cm}
	\newline
	\begin{subtable}[h]{\textwidth}
	 \scriptsize
		\centering
		\resizebox{0.65\columnwidth}{!}{
		\begin{tabular}{c c c c c c c }\toprule
			\multicolumn{7}{c}{Validation datasets}                            \\
			\cmidrule(lr){1-7}
			ADE20K & BDD100K & Cityscapes & COCO & IDD  & Mapillary & SUN-RGBD \\
			\cmidrule(lr){2-2}\cmidrule(lr){1-1}\cmidrule(lr){3-3}\cmidrule(lr){4-4}\cmidrule(lr){5-5}\cmidrule(lr){6-6}\cmidrule(lr){7-7}
			0.0    & 0.0     & 0.0        & 0.0  & 0.0  & 0.0       & 0.0      \\
			0.0    & 16.5    & 19.5       & 0.0  & 14.3 & 2.7       & 5.9      \\
			0.0    & 15.2    & 0.0        & 3.4  & 2.6  & 0.0       & 28.8     \\
			2.1    & 0.9     & 19.5       & 0.0  & 16.9 & 0.0       & 34.7     \\
			1.0    & 2.2     & 43.6       & 3.4  & 4.0  & 2.7       & 0.0      \\
			3.1    & 2.6     & 0.0        & 0.0  & 4.0  & 21.3      & 0.0      \\
			0.0    & 0.0     & 43.6       & 0.0  & 0.0  & 21.3      & 0.0      \\\cmidrule(lr){1-7}
			\multicolumn{7}{c}{Average absolute mIoU difference: 6.85}         \\\bottomrule
		\end{tabular}
		}
		\caption{Differences between the mIoUs of the reference ranking (\cref{tab:mseg_ref_val}) and the mIoUs for the predicted \texttt{dna-emd} ranking (\cref{tab:mseg_val_miou}).}
		\label{tab:mseg_val_diffs}
	\end{subtable}
	\newline
	\caption{Detailed results comparing the observed cross-dataset generalisation of the HRNet-W48 semantic segmentation models to predictions relying only on datasets using \textbf{\texttt{dna-emd}} with the \textbf{HRNet-W48 feature extractor trained on validation domains}. The final value reported in \cref{tab:mseg-summary} corresponds to the average of the values in \cref{tab:mseg_val_diffs}.}
	\label{tab:detailed-mseg-val}
\end{table*}

\begin{table*}[ht]
	\centering
	\begin{subtable}[h]{\textwidth}
		\centering
		\resizebox{\columnwidth}{!}{
			\begin{tabular}{l  l  l  l  l  l  l  l}\toprule
				\multicolumn{1}{c}{}                                                        & \multicolumn{7}{c}{Validation datasets}                                                                                                                       \\
				\cmidrule(lr){2-8}
				                                                                            & ADE20K                                  & BDD100K           & Cityscapes        & COCO             & IDD               & Mapillary         & SUN-RGBD         \\
				\cmidrule(lr){2-2}\cmidrule(lr){3-3}\cmidrule(lr){4-4}\cmidrule(lr){5-5}\cmidrule(lr){6-6}\cmidrule(lr){7-7}\cmidrule(lr){8-8}
				\multirow{7}{*}{\rotatebox{90}{\parbox{3cm}{\centering Training datasets}}} & 45.3 - ADE20K                           & 63.2 - BDD100K    & 77.6 - Cityscapes & 52.6 - COCO      & 64.8 - IDD        & 56.2 - Mapillary  & 43.9 - SUN-RGBD  \\
				                                                                            & 19.6 - COCO                             & 60.2 - Mapillary  & 69.7 - Mapillary  & 14.5 - ADE20K    & 48.2 - Mapillary  & 26.7 - COCO       & 35.3 - ADE20K    \\
				                                                                            & 7.1 - SUN-RGBD                          & 45.0 - Cityscapes & 60.9 - BDD100K    & 6.7 - Mapillary  & 33.9 - BDD100K    & 24.3 - ADE20K     & 29.4 - COCO      \\
				                                                                            & 6.2 - Mapillary                         & 44.1 - COCO       & 50.2 - IDD        & 3.7 - BDD100K    & 31.3 - Cityscapes & 24.3 - IDD        & 0.6 - IDD        \\
				                                                                            & 4.1 - BDD100K                           & 43.7 - IDD        & 46.2 - COCO       & 3.3 - SUN-RGBD   & 31.0 - COCO       & 24.0 - BDD100K    & 0.2 - BDD100K    \\
				                                                                            & 3.1 - Cityscapes                        & 41.5 - ADE20K     & 44.3 - ADE20K     & 3.1 - Cityscapes & 27.0 - ADE20K     & 22.4 - Cityscapes & 0.2 - Cityscapes \\
				                                                                            & 3.1 - IDD                               & 2.2 - SUN-RGBD    & 2.6 - SUN-RGBD    & 3.1 - IDD        & 1.0 - SUN-RGBD    & 1.1 - SUN-RGBD    & 0.2 - Mapillary  \\\bottomrule
			\end{tabular}}
		\caption{Observed cross-dataset generalisation on semantic segmentation from Lambert \etal~\cite{lambertMSegCompositeDataset2020} (mIoU). Each column corresponds to the evaluation of one dataset (validation set). Rows are ordered by a cross-generalisation performance from training on each dataset (training set).}
		\label{tab:mseg_ref_nfd}
	\end{subtable}
	\newline
	\vspace*{0.4cm}
	\newline
	\begin{subtable}[h]{\textwidth}
		\centering
		\resizebox{\columnwidth}{!}{
			\begin{tabular}{l l l l l l l l }\toprule
				\multicolumn{1}{c}{}                                                          & \multicolumn{7}{c}{Validation datasets}                                                                                                                        \\
				\cmidrule(lr){2-8}
				                                                                              & ADE20K                                  & BDD100K           & Cityscapes        & COCO              & IDD              & Mapillary         & SUN-RGBD          \\
				\cmidrule(lr){2-2}\cmidrule(lr){3-3}\cmidrule(lr){4-4}\cmidrule(lr){5-5}\cmidrule(lr){6-6}\cmidrule(lr){7-7}\cmidrule(lr){8-8}
				\multirow{7}{*}{\rotatebox{90}{\parbox{2.9cm}{\centering Training datasets}}} & 0.0 - ADE20K                            & 0.01 - BDD100K    & 0.01 - Cityscapes & 0.0 - COCO        & 0.01 - IDD       & 0.0 - Mapillary   & 0.0 - SUN-RGBD    \\
				                                                                              & 0.11 - COCO                             & 0.06 - Mapillary  & 0.13 - Mapillary  & 0.12 - ADE20K     & 0.09 - BDD100K   & 0.07 - BDD100K    & 0.16 - ADE20K     \\
				                                                                              & 0.15 - SUN-RGBD                         & 0.08 - IDD        & 0.16 - IDD        & 0.23 - SUN-RGBD   & 0.09 - Mapillary & 0.07 - IDD        & 0.24 - COCO       \\
				                                                                              & 0.32 - IDD                              & 0.17 - Cityscapes & 0.17 - BDD100K    & 0.31 - Mapillary  & 0.2 - Cityscapes & 0.14 - Cityscapes & 0.49 - IDD        \\
				                                                                              & 0.32 - BDD100K                          & 0.32 - ADE20K     & 0.43 - COCO       & 0.34 - IDD        & 0.34 - ADE20K    & 0.31 - COCO       & 0.51 - Mapillary  \\
				                                                                              & 0.32 - Mapillary                        & 0.39 - COCO       & 0.47 - ADE20K     & 0.39 - BDD100K    & 0.37 - COCO      & 0.33 - ADE20K     & 0.52 - BDD100K    \\
				                                                                              & 0.48 - Cityscapes                       & 0.52 - SUN-RGBD   & 0.6 - SUN-RGBD    & 0.44 - Cityscapes & 0.51 - SUN-RGBD  & 0.5 - SUN-RGBD    & 0.62 - Cityscapes \\\bottomrule
			\end{tabular}}
		\caption{Ordered datasets ranked by \texttt{dna-fd} with corresponding FD values. The FD here is computed using the last layer of the Mugs (ViT-B/16) feature extractor.}
		\label{tab:mseg_ours_dists_nfd}
	\end{subtable}
	\newline
	\vspace*{0.4 cm}
	\newline
	\begin{subtable}[h]{\textwidth}
		\centering
		\resizebox{\columnwidth}{!}{
			\begin{tabular}{l l l l l l l l }\toprule
				\multicolumn{1}{c}{}                                                          & \multicolumn{7}{c}{Validation datasets}                                                                                                                       \\
				\cmidrule(lr){2-8}
				                                                                              & ADE20K                                  & BDD100K           & Cityscapes        & COCO             & IDD               & Mapillary         & SUN-RGBD         \\
				\cmidrule(lr){2-2}\cmidrule(lr){3-3}\cmidrule(lr){4-4}\cmidrule(lr){5-5}\cmidrule(lr){6-6}\cmidrule(lr){7-7}\cmidrule(lr){8-8}
				\multirow{7}{*}{\rotatebox{90}{\parbox{2.9cm}{\centering Training datasets}}} & 45.3 - ADE20K                           & 63.2 - BDD100K    & 77.6 - Cityscapes & 52.6 - COCO      & 64.8 - IDD        & 56.2 - Mapillary  & 43.9 - SUN-RGBD  \\
				                                                                              & 19.6 - COCO                             & 60.2 - Mapillary  & 69.7 - Mapillary  & 14.5 - ADE20K    & 33.9 - BDD100K    & 24.0 - BDD100K    & 35.3 - ADE20K    \\
				                                                                              & 7.1 - SUN-RGBD                          & 43.7 - IDD        & 50.2 - IDD        & 3.3 - SUN-RGBD   & 48.2 - Mapillary  & 24.3 - IDD        & 29.4 - COCO      \\
				                                                                              & 3.1 - IDD                               & 45.0 - Cityscapes & 60.9 - BDD100K    & 6.7 - Mapillary  & 31.3 - Cityscapes & 22.4 - Cityscapes & 0.6 - IDD        \\
				                                                                              & 4.1 - BDD100K                           & 41.5 - ADE20K     & 46.2 - COCO       & 3.1 - IDD        & 27.0 - ADE20K     & 26.7 - COCO       & 0.2 - Mapillary  \\
				                                                                              & 6.2 - Mapillary                         & 44.1 - COCO       & 44.3 - ADE20K     & 3.7 - BDD100K    & 31.0 - COCO       & 24.3 - ADE20K     & 0.2 - BDD100K    \\
				                                                                              & 3.1 - Cityscapes                        & 2.2 - SUN-RGBD    & 2.6 - SUN-RGBD    & 3.1 - Cityscapes & 1.0 - SUN-RGBD    & 1.1 - SUN-RGBD    & 0.2 - Cityscapes \\\bottomrule
			\end{tabular}}
		\caption{Ordered datasets ranked by \texttt{dna-fd} with corresponding mIoU values. The ranking is taken from \cref{tab:mseg_ours_dists_nfd}, but we show the mIoUs for the corresponding datasets from \cref{tab:mseg_ref_nfd} instead.}
		\label{tab:mseg_ours_miou_nfd}
	\end{subtable}
	\newline
	\vspace*{0.4 cm}
	\newline
	\begin{subtable}[h]{\textwidth}
	 \scriptsize
		\centering
		\resizebox{0.65\columnwidth}{!}{
		\begin{tabular}{c c c c c c c }\toprule
			\multicolumn{7}{c}{Validation datasets}                            \\
			\cmidrule(lr){1-7}
			ADE20K & BDD100K & Cityscapes & COCO & IDD  & Mapillary & SUN-RGBD \\
			\cmidrule(lr){1-1}\cmidrule(lr){2-2}\cmidrule(lr){3-3}\cmidrule(lr){4-4}\cmidrule(lr){5-5}\cmidrule(lr){6-6}\cmidrule(lr){7-7}
			0.0    & 0.0     & 0.0        & 0.0  & 0.0  & 0.0       & 0.0      \\
			0.0    & 0.0     & 0.0        & 0.0  & 14.3 & 2.7       & 0.0      \\
			0.0    & 1.3     & 10.7       & 3.4  & 14.3 & 0.0       & 0.0      \\
			3.1    & 0.9     & 10.7       & 3.0  & 0.0  & 1.9       & 0.0      \\
			0.0    & 2.2     & 0.0        & 0.2  & 4.0  & 2.7       & 0.0      \\
			3.1    & 2.6     & 0.0        & 0.6  & 4.0  & 1.9       & 0.0      \\
			0.0    & 0.0     & 0.0        & 0.0  & 0.0  & 0.0       & 0.0      \\\cmidrule(lr){1-7}
			\multicolumn{7}{c}{Average absolute mIoU difference: 1.79}         \\\bottomrule
		\end{tabular}
		}
		\caption{Differences between the mIoUs of the reference ranking (\cref{tab:mseg_ref_nfd}) and the mIoUs for the predicted \texttt{dna-fd} ranking (\cref{tab:mseg_ours_miou_nfd}).}
		\label{tab:mseg_diffs_nfd}
	\end{subtable}
	\newline
	\caption{Detailed results comparing the observed cross-dataset generalisation of the HRNet-W48 semantic segmentation models to predictions relying only on datasets using \textbf{\texttt{dna-fd}} with the \textbf{Mugs (ViT-B/16) feature extractor}. The final value reported in \cref{tab:mseg-summary} corresponds to the average of the values in \cref{tab:mseg_diffs_nfd}.}
	\label{tab:detailed-mseg-nfd}
\end{table*}

\begin{table*}[ht]
	\centering
	\begin{subtable}[h]{\textwidth}
		\centering
		\resizebox{\columnwidth}{!}{
			\begin{tabular}{l  l  l  l  l  l  l  l}\toprule
				\multicolumn{1}{c}{}                                                        & \multicolumn{7}{c}{Validation datasets}                                                                                                                       \\
				\cmidrule(lr){2-8}
				                                                                            & ADE20K                                  & BDD100K           & Cityscapes        & COCO             & IDD               & Mapillary         & SUN-RGBD         \\
				\cmidrule(lr){2-2}\cmidrule(lr){3-3}\cmidrule(lr){4-4}\cmidrule(lr){5-5}\cmidrule(lr){6-6}\cmidrule(lr){7-7}\cmidrule(lr){8-8}
				\multirow{7}{*}{\rotatebox{90}{\parbox{3cm}{\centering Training datasets}}} & 45.3 - ADE20K                           & 63.2 - BDD100K    & 77.6 - Cityscapes & 52.6 - COCO      & 64.8 - IDD        & 56.2 - Mapillary  & 43.9 - SUN-RGBD  \\
				                                                                            & 19.6 - COCO                             & 60.2 - Mapillary  & 69.7 - Mapillary  & 14.5 - ADE20K    & 48.2 - Mapillary  & 26.7 - COCO       & 35.3 - ADE20K    \\
				                                                                            & 7.1 - SUN-RGBD                          & 45.0 - Cityscapes & 60.9 - BDD100K    & 6.7 - Mapillary  & 33.9 - BDD100K    & 24.3 - ADE20K     & 29.4 - COCO      \\
				                                                                            & 6.2 - Mapillary                         & 44.1 - COCO       & 50.2 - IDD        & 3.7 - BDD100K    & 31.3 - Cityscapes & 24.3 - IDD        & 0.6 - IDD        \\
				                                                                            & 4.1 - BDD100K                           & 43.7 - IDD        & 46.2 - COCO       & 3.3 - SUN-RGBD   & 31.0 - COCO       & 24.0 - BDD100K    & 0.2 - BDD100K    \\
				                                                                            & 3.1 - Cityscapes                        & 41.5 - ADE20K     & 44.3 - ADE20K     & 3.1 - Cityscapes & 27.0 - ADE20K     & 22.4 - Cityscapes & 0.2 - Cityscapes \\
				                                                                            & 3.1 - IDD                               & 2.2 - SUN-RGBD    & 2.6 - SUN-RGBD    & 3.1 - IDD        & 1.0 - SUN-RGBD    & 1.1 - SUN-RGBD    & 0.2 - Mapillary  \\\bottomrule
			\end{tabular}}
		\caption{Observed cross-dataset generalisation on semantic segmentation from Lambert \etal~\cite{lambertMSegCompositeDataset2020} (mIoU). Each column corresponds to the evaluation of one dataset (validation set). Rows are ordered by a cross-generalisation performance from training on each dataset (training set).}
		\label{tab:mseg_ref_fd}
	\end{subtable}
	\newline
	\vspace*{0.4cm}
	\newline
	\begin{subtable}[h]{\textwidth}
		\centering
		\resizebox{\columnwidth}{!}{
			\begin{tabular}{l l l l l l l l }\toprule
				\multicolumn{1}{c}{}                                                          & \multicolumn{7}{c}{Validation datasets}                                                                                                                                    \\
				\cmidrule(lr){2-8}
				                                                                              & ADE20K                                  & BDD100K             & Cityscapes         & COCO                & IDD                 & Mapillary           & SUN-RGBD            \\
				\cmidrule(lr){2-2}\cmidrule(lr){3-3}\cmidrule(lr){4-4}\cmidrule(lr){5-5}\cmidrule(lr){6-6}\cmidrule(lr){7-7}\cmidrule(lr){8-8}
				\multirow{7}{*}{\rotatebox{90}{\parbox{2.9cm}{\centering Training datasets}}} & 19.37 - ADE20K                          & 31.16 - BDD100K     & 40.74 - Cityscapes & 9.68 - COCO         & 54.03 - IDD         & 9.78 - Mapillary    & 71.23 - SUN-RGBD    \\
				                                                                              & 233.41 - COCO                           & 101.67 - Mapillary  & 196.74 - Mapillary & 227.36 - ADE20K     & 160.51 - BDD100K    & 96.19 - BDD100K     & 332.54 - ADE20K     \\
				                                                                              & 279.51 - SUN-RGBD                       & 141.84 - IDD        & 225.68 - IDD       & 389.04 - SUN-RGBD   & 161.9 - Mapillary   & 123.14 - IDD        & 460.08 - COCO       \\
				                                                                              & 398.31 - Mapillary                      & 215.55 - Cityscapes & 232.53 - BDD100K   & 417.86 - Mapillary  & 255.92 - Cityscapes & 171.68 - Cityscapes & 638.82 - IDD        \\
				                                                                              & 401.76 - BDD100K                        & 423.12 - ADE20K     & 543.35 - COCO      & 457.03 - IDD        & 470.54 - ADE20K     & 401.26 - ADE20K     & 652.68 - BDD100K    \\
				                                                                              & 416.94 - IDD                            & 508.35 - COCO       & 550.05 - ADE20K    & 476.86 - BDD100K    & 519.04 - COCO       & 428.07 - COCO       & 660.09 - Mapillary  \\
				                                                                              & 522.29 - Cityscapes                     & 645.3 - SUN-RGBD    & 701.34 - SUN-RGBD  & 509.45 - Cityscapes & 656.16 - SUN-RGBD   & 630.26 - SUN-RGBD   & 704.86 - Cityscapes \\\bottomrule
			\end{tabular}}
		\caption{Ordered datasets ranked by \texttt{fd} with corresponding FD values. The FD here is computed using the last layer of the Mugs (ViT-B/16) feature extractor.}
		\label{tab:mseg_ours_dists_fd}
	\end{subtable}
	\newline
	\vspace*{0.4 cm}
	\newline
	\begin{subtable}[h]{\textwidth}
		\centering
		\resizebox{\columnwidth}{!}{
			\begin{tabular}{l l l l l l l l }\toprule
				\multicolumn{1}{c}{}                                                          & \multicolumn{7}{c}{Validation datasets}                                                                                                                       \\
				\cmidrule(lr){2-8}
				                                                                              & ADE20K                                  & BDD100K           & Cityscapes        & COCO             & IDD               & Mapillary         & SUN-RGBD         \\
				\cmidrule(lr){2-2}\cmidrule(lr){3-3}\cmidrule(lr){4-4}\cmidrule(lr){5-5}\cmidrule(lr){6-6}\cmidrule(lr){7-7}\cmidrule(lr){8-8}
				\multirow{7}{*}{\rotatebox{90}{\parbox{2.9cm}{\centering Training datasets}}} & 45.3 - ADE20K                           & 63.2 - BDD100K    & 77.6 - Cityscapes & 52.6 - COCO      & 64.8 - IDD        & 56.2 - Mapillary  & 43.9 - SUN-RGBD  \\
				                                                                              & 19.6 - COCO                             & 60.2 - Mapillary  & 69.7 - Mapillary  & 14.5 - ADE20K    & 33.9 - BDD100K    & 24.0 - BDD100K    & 35.3 - ADE20K    \\
				                                                                              & 7.1 - SUN-RGBD                          & 43.7 - IDD        & 50.2 - IDD        & 3.3 - SUN-RGBD   & 48.2 - Mapillary  & 24.3 - IDD        & 29.4 - COCO      \\
				                                                                              & 6.2 - Mapillary                         & 45.0 - Cityscapes & 60.9 - BDD100K    & 6.7 - Mapillary  & 31.3 - Cityscapes & 22.4 - Cityscapes & 0.6 - IDD        \\
				                                                                              & 4.1 - BDD100K                           & 41.5 - ADE20K     & 46.2 - COCO       & 3.1 - IDD        & 27.0 - ADE20K     & 24.3 - ADE20K     & 0.2 - BDD100K    \\
				                                                                              & 3.1 - IDD                               & 44.1 - COCO       & 44.3 - ADE20K     & 3.7 - BDD100K    & 31.0 - COCO       & 26.7 - COCO       & 0.2 - Mapillary  \\
				                                                                              & 3.1 - Cityscapes                        & 2.2 - SUN-RGBD    & 2.6 - SUN-RGBD    & 3.1 - Cityscapes & 1.0 - SUN-RGBD    & 1.1 - SUN-RGBD    & 0.2 - Cityscapes \\\bottomrule
			\end{tabular}}
		\caption{Ordered datasets ranked by \texttt{fd} with corresponding mIoU values. The ranking is taken from \cref{tab:mseg_ours_dists_fd}, but we show the mIoUs for the corresponding datasets from \cref{tab:mseg_ref_fd} instead.}
		\label{tab:mseg_ours_miou_fd}
	\end{subtable}
	\newline
	\vspace*{0.4 cm}
	\newline
	\begin{subtable}[h]{\textwidth}
	 \scriptsize
		\centering
		\resizebox{0.65\columnwidth}{!}{
		\begin{tabular}{c c c c c c c }\toprule
			\multicolumn{7}{c}{Validation datasets}                            \\
			\cmidrule(lr){1-7}
			ADE20K & BDD100K & Cityscapes & COCO & IDD  & Mapillary & SUN-RGBD \\
			\cmidrule(lr){1-1}\cmidrule(lr){2-2}\cmidrule(lr){3-3}\cmidrule(lr){4-4}\cmidrule(lr){5-5}\cmidrule(lr){6-6}\cmidrule(lr){7-7}
			0.0    & 0.0     & 0.0        & 0.0  & 0.0  & 0.0       & 0.0      \\
			0.0    & 0.0     & 0.0        & 0.0  & 14.3 & 2.7       & 0.0      \\
			0.0    & 1.3     & 10.7       & 3.4  & 14.3 & 0.0       & 0.0      \\
			0.0    & 0.9     & 10.7       & 3.0  & 0.0  & 1.9       & 0.0      \\
			0.0    & 2.2     & 0.0        & 0.2  & 4.0  & 0.3       & 0.0      \\
			0.0    & 2.6     & 0.0        & 0.6  & 4.0  & 4.3       & 0.0      \\
			0.0    & 0.0     & 0.0        & 0.0  & 0.0  & 0.0       & 0.0      \\\cmidrule(lr){1-7}
			\multicolumn{7}{c}{Average absolute mIoU difference: 1.66}         \\\bottomrule
		\end{tabular}
		}
		\caption{Differences between the mIoUs of the reference ranking (\cref{tab:mseg_ref_fd}) and the mIoUs for the predicted \texttt{fd} ranking (\cref{tab:mseg_ours_miou_fd}).}
		\label{tab:mseg_diffs_fd}
	\end{subtable}
	\newline
	\caption{Detailed results comparing the observed cross-dataset generalisation of the HRNet-W48 semantic segmentation models to predictions relying only on datasets using \textbf{\texttt{fd}} with the \textbf{Mugs (ViT-B/16) feature extractor}. The final value reported in \cref{tab:mseg-summary} corresponds to the average of the values in \cref{tab:mseg_diffs_fd}.}
	\label{tab:detailed-mseg-fd}
\end{table*}

\end{document}